# Using Pivot Consistency to Decompose and Solve Functional CSPs

**Philippe David** DAVID@LIRMM.FR
*LIRMM (UMR C9928 CNRS/université Montpellier II)*
*161 rue Ada, 34392 Montpellier Cedex 5 — France*

## Abstract

Many studies have been carried out in order to increase the search efficiency of constraint satisfaction problems; among them, some make use of *structural* properties of the constraint network; others take into account *semantic* properties of the constraints, generally assuming that *all* the constraints possess the given property. In this paper, we propose a new decomposition method benefiting from both semantic properties of *functional* constraints (not *bijective* constraints) and structural properties of the network; furthermore, not all the constraints need to be functional. We show that under some conditions, the existence of solutions can be guaranteed. We first characterize a particular subset of the variables, which we name a *root set*. We then introduce *pivot consistency*, a new local consistency which is a weak form of path consistency and can be achieved in $O(n^2d^2)$ complexity (instead of $O(n^3d^3)$ for path consistency), and we present associated properties; in particular, we show that any consistent instantiation of the root set can be linearly extended to a solution, which leads to the presentation of the aforementioned new method for solving by decomposing functional CSPs.

## 1. Introduction

Improving search efficiency is one of the main goals of artificial intelligence. Efficiency is particularly important for researchers interested in constraint satisfaction problems (CSPs), also called *constraint networks*. In order to achieve this aim, various approaches have been tried, such as filtering techniques (mainly arc and path consistencies) (Montanari, 1974; Mackworth, 1977; Mohr & Henderson, 1986; Bessière, 1994) or improvements to the backtrack process (Haralick & Elliot, 1980; Dechter & Pearl, 1988; Ginsberg, 1993; Prosser, 1993). Other work concerned the characterization of classes of polynomial problems, based on the size of their domains (Dechter, 1992) or on the structure of the constraint network (Freuder, 1978), leading to the presentation of decomposition methods such as the *cycle-cutset* (Dechter, 1990) or the *tree clustering* (Dechter & Pearl, 1989) methods. A more recent approach consists of taking into account *semantic* properties of the constraints (as opposed to *structural* or *topological* properties of the network) to achieve arc consistency efficiently for specific classes of constraints (Van Hentenryck, Deville, & Teng, 1992; Mohr & Masini, 1988), or to characterize some tractable classes of problems (van Beek, 1992; van Beek & Dechter, 1994; Kirousis, 1993; Cooper, Cohen, & Jeavons, 1994). We present these results further in this paper.

Some frequently encountered constraints are *functional* constraints, for instance in peptide synthesis (Janssen, Jégou, Nouguier, Vilarem, & Castro, 1990) or in Constraint Logic

Programming (Van Hentenryck et al., 1992). Many constraint-based systems and languages use functional constraints, including THINGLAB (Borning, 1981), GARNET (Myers, Giuse, & Vander Zanden, 1992) and KALEIDOSCOPE (Freeman-Benson & Borning, 1992).

In this paper, we study functional constraints: the relations they represent are partial functions (notice those are *not* the same as *bijective* constraints: see sections 2.1 and 2.2). More precisely, we study the properties given to them by a new local consistency, *pivot consistency*. We notably show that, under some conditions, this local consistency guarantees the existence of solutions and makes them easier to find: first, we characterize a subset of the variables, called a *root set* and denoted $\mathcal{R}$; we then show that if the network is pivot consistent, then any consistent instantiation of $\mathcal{R}$ can be linearly extended to a solution, provided the instantiation order is $\mathcal{R}$-*compatible* (*i.e.*, it possesses some topological properties associated with $\mathcal{R}$ and the functional constraints).

We then introduce a new method for solving any functional CSP by decomposing it into two subproblems: first, finding a consistent instantiation of the root set, and second, extending this partial instantiation to a solution (in a backtrack-free manner).

The main difficulty is clearly to find a consistent instantiation of the root set, which remains exponential in its size. The method we present here is therefore all the more efficient since the root set is of small size.

Another aspect we wish to point out is that, unlike most of the work dealing with constraints which possess some specific properties, this method not only applies when *all* the constraints possess the given property (or properties), but also when only *some* of them do.

This paper is organized as follows: section 2 first introduces the basic definitions and notations regarding CSPs which will be used in the following. We then define functional constraints and discuss previous work and its relations with this paper. The last part of the section presents notions associated with functional constraints, which are mainly found in graph theory. We then introduce pivot consistency and an $O(n^2d^2)$ algorithm allowing to achieve the latter in section 3, before presenting in section 4 the properties supplied to a CSP by this consistency. We finally propose a method based on those properties for solving CSPs composed of functional constraints (with or without additional non-functional constraints).

## 2. Preliminaries

**Definition 2.1** *A binary* CSP $\mathcal{P}$ *is defined by* $(X, D, C, R)$*, where*

- $X = \{x_1, x_2, \dots, x_i, \dots, x_n\}$ *is the set of its* $n$ *variables.*
- $D = \{D_1, \dots, D_i, \dots, D_n\}$ *is the set of its* $n$ *domains, where each* $D_i = \{a_i, b_i, \dots\}$ *is the set of the possible values for variable* $x_i$*; the size of the largest domain of* $D$ *is called* $d$*.*
- $C$ *is the set of its* $e$ *constraints* $\{c_1, \dots, c_k, \dots, c_e\}$*; for the sake of simplicity, a constraint* $c_k = \{x_i, x_j\}$ *involving variables* $x_i$ *and* $x_j$ *is denoted* $c_{ij}$*.*
- $R$ *is the set of the* $e$ *relations associated with the constraints, where the relation* $R_{ij}$ *is the subset of the Cartesian product* $D_i \times D_j$ *specifying the pairs of mutually compatible values.*

A graph $\mathcal{G} = (X, C)$ called the *constraint graph* can be associated with $\mathcal{P}$, where the vertices and the edges (or arcs) respectively represent the variables and the constraints. Another graph, representing the relations of the CSP, can also be defined: the *consistency graph.* Figure 1 represents the constraint and consistency graphs of the following example:

> *A travel agency offers five destinations to its customers: Paris, London, Washington, New-York and Madrid, and employs three guides to accompany them: Alice, Bob and Chris. These guides provide the information below:*
>
> - *Alice wishes to go to Paris or New-York, but she only speaks French.*
> - *Bob speaks English and French, and does not want to go to Madrid (but accepts any other city).*
> - *Chris only speaks Spanish, and refuses to go to any city but New-York.*
>
> *The manager would like to find all the possibilities left to him, i.e., which guide he can send to which city, with which currency, knowing that each guide must of course speak the language of the country he (or she) visits, and the currency must be that of the country.*

This problem can be encoded by the CSP $\mathcal{P}_{ex}$, composed of five variables and five constraints:

- The set of the five variables is
  $X = \{$*GUIDES, CITIES, COUNTRIES, CURRENCIES, LANGUAGES*$\}$

- Their five domains are:
  $D_{GUIDES} = \{$*Alice (A), Bob (B), Chris (C)*$\}$
  $D_{CITIES} = \{$*Paris, London, Washington, New-York, Madrid*$\}$
  $D_{COUNTRIES} = \{$*France, GB, USA, Spain*$\}$
  $D_{CURRENCIES} = \{$*FrF, £, \$, Pes*$\}$
  $D_{LANGUAGES} = \{$*French, English, Spanish*$\}$

- The set of the five constraints is:
  $C = \{C_{GUIDES\text{–}CITIES}, C_{CITIES\text{–}COUNTRIES}, C_{GUIDES\text{–}LANGUAGES}, C_{COUNTRIES\text{–}CURRENCIES}, C_{COUNTRIES\text{–}LANGUAGES}\}$

  where these constraints respectively represent the cities the guides wish (or accept) to visit, the countries the cities belong to, the languages spoken by the guides, the currencies used in the countries, and the official languages of the countries. Notice there is no explicit constraint between the currencies and the guides: this constraint will be induced by the cities (and thus the countries) visited by the guides on the one hand, and by the currencies of the countries on the other hand.

- The five associated relations are:

  $R_{GUIDES\text{–}CITIES} = \{$*(Alice,Paris), (Alice,New-York), (Bob,Paris), (Bob,London), (Bob,Washington), (Bob,New-York), (Chris,New-York)*$\}$

  $R_{GUIDES\text{–}LANGUAGES} = \{$*(Alice,French), (Bob,French), (Bob,English), (Chris,Spanish)*$\}$

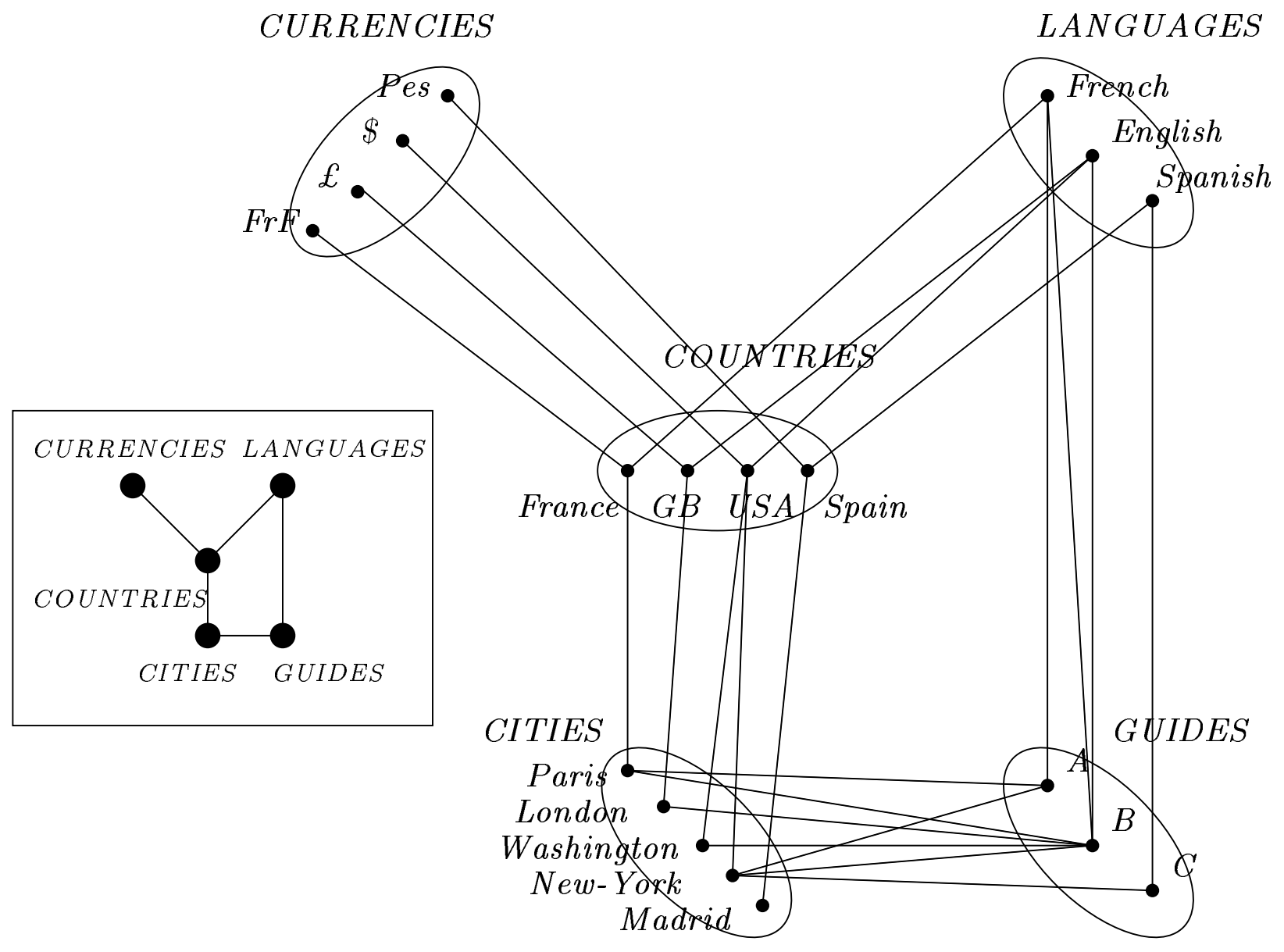

Figure 1: Constraint graph and Consistency graph of $\mathcal{P}_{ex}$

$$R_{\textit{CITIES–COUNTRIES}} = \{\textit{(Paris,France), (London,GB), (Washington,USA),}$$
$$\textit{(New-York,USA), (Madrid,Spain)}\}$$

$$R_{\textit{COUNTRIES–CURRENCIES}} = \{\textit{(France,FrF), (GB,£), (USA,\$), (Spain,Pes)}\}$$

$$R_{\textit{COUNTRIES–LANGUAGES}} = \{\textit{(France,French), (GB,English), (USA,English),}$$
$$\textit{(Spain,Spanish)}\}$$

If there is no constraint between variables $X_i$ and $X_j$ ($C_{ij} \notin C$), all pairs of values are therefore allowed: $R_{ij} = D_i \times D_j$. $R_{ij}$ is then called a *universal relation.*

The notion of *support* is related to the pairs that belong or not to the relations: we say that a value $a_k \in D_k$ is a *support* of (or supports) a value $a_i \in D_i$ for the constraint $C_{ik}$ if and only if $(a_i, a_k) \in R_{ik}$. Also, the value $a_k \in D_k$ is a support of the pair $(a_i, a_j) \in R_{ij}$ if and only if $a_k$ both supports $a_i$ for the constraint $C_{ik}$ and $a_j$ for the constraint $C_{jk}$.

$Y$ stands for any subset of $X$, and $Y_k$ represents the set of the first $k$ variables of $X$, w.r.t. an order explained in the context. The default order will be $X_1, X_2, \ldots, X_{n-1}, X_n$. Also, the order for $\mathcal{P}_{ex}$ will be *GUIDES, CITIES, COUNTRIES, CURRENCIES, LANGUAGES.*

Given $Y_t = \{X_1, \ldots, X_t\}$ a subset of $X$, the assignment to each variable $X_i$ of $Y_t$ of a value $a_i$ of $D_i$ is called a *(partial) instantiation of* $Y_t$. Such an instantiation $I_t = (a_1, \ldots, a_i, \ldots, a_j, \ldots, a_t)$ is consistent if and only if every constraint included in $Y_t$ is satisfied; formally, $\forall i, j \leq t$ s.t. $C_{ij} \in C$, $(a_i, a_j) \in R_{ij}$. For instance, neither *(Alice, London)* nor *(Bob, Paris, Spain)* are consistent, whereas *(Alice, Paris)* and *(Chris, New-York, Peseta, Spanish)* both are. A *solution* is therefore defined as a consistent instantiation of $X$. A solution to $X$ is thus *(Alice, Paris, France, FrF, French).*

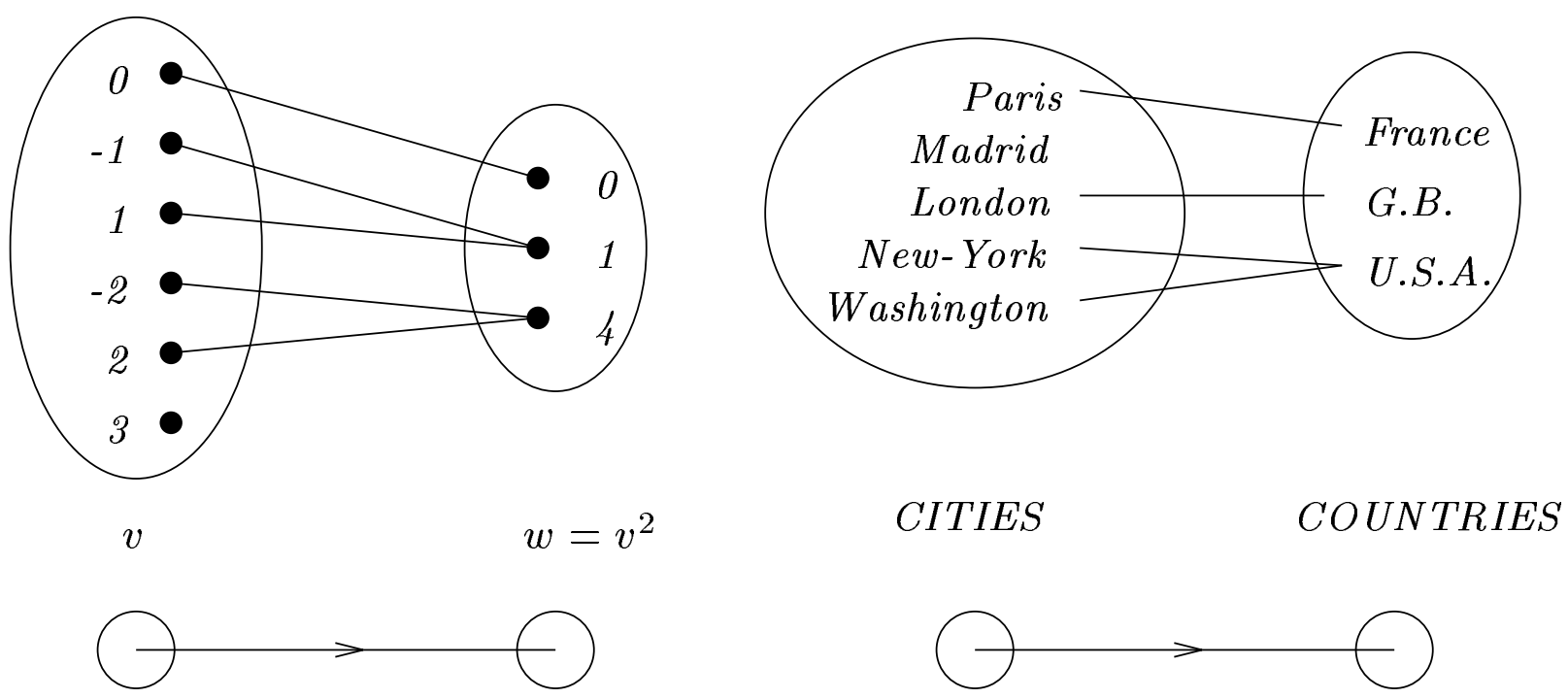

Figure 2: Two examples of functional constraints

### 2.1 Functional Constraints: Definition

We first define what we call *functional constraints* (Figure 2 presents two examples of such constraints), before discussing related work. We then recall some graph theory notions before characterizing a property necessary for the method we present in this paper.

**Definition 2.2** *Given two variables* $X_i$ *and* $X_k$*, we denote* $X_i \rightarrow X_k$ *iff for all* $a_i$ *in* $D_i$*, there is at most one* $a_k$ *in* $D_k$ *such that* $(a_i, a_k) \in R_{ik}$*. We then denote* $a_k = f_{ik}(a_i)$ *(or* $a_k = f_{i \rightarrow k}(a_i)$*) if this value exists,* $\varepsilon$ *otherwise.*

*A constraint* $C_{ik}$ *is* functional *iff* $X_i \rightarrow X_k$ *or* $X_k \rightarrow X_i$*. In the following, we will call* $X_i$ *the* origin *and* $X_k$ *the* target *of the functional constraint* $X_i \rightarrow X_k$*.*

From this definition, we can deduce a partition of the constraint set of the CSP: on the one hand $C_f$, the set of the functional constraints, on the other hand $C_o$, the others (the non functional constraints). A CSP is said to be functional if it contains *some* functional constraints ($C_f \neq \emptyset$), and *strictly functional* if it *only* contains such constraints: $C = C_f$ ($C_o = \emptyset$).

### 2.2 Related Work

In the field of constraint satisfaction problems, one of the most important objectives is to reduce computation time. Many studies have been carried out to achieve this (see for instance the references in the introduction of this paper). They permit to reduce, sometimes drastically, the time required to solve CSPs. However, an obstacle limits their efficiency: they are general methods, and, as such, they do not take into account the specificities of the problems to be solved, notably w.r.t. *semantics* of the constraints. This deficiency was overcome by various (and, for most of them, recent) results. They exhibit some classes of constraints which possess particular properties. These results fall into two classes: on the one hand, those proposing algorithms specifically adapted to some classes of constraints, more efficient than general algorithms, and on the other hand, those characterizing classes of "tractable" problems. In this section as in the rest of this paper, we will suppose that, unless otherwise stated, the networks are binary.

The first class contains those techniques which make use of the *semantics* of the constraints to propose specific processings, namely arc consistency filterings.

- Mohr and Masini (1988) propose to modify AC-4 to improve its efficiency on *equality, inequality* and *disequality* constraints. They reduce time complexity of AC-4 from $O(ed^2)$ in general case, to $O(ed)$ for those specific constraints.

- AC-5, a new arc-consistency algorithm, is presented by Van Hentenryck *et al.* (1992). Actually, AC-5 is not an algorithm: it should rather be considered as a model with several possible specializations, according to the classes of constraints to process (*e.g., functional*[1], *anti-functional* and *monotonic* constraints — notice that these constraints respectively correspond to Mohr and Masini's *equality, disequality* and *inequality* constraints). AC-5 achieves arc-consistency in $O(ed)$ for those constraints, which is consistent with the former result.[2]

The second class contains those techniques which identify constraints for which some local consistency is sufficient to guarantee global consistency, or propose polynomial algorithms to solve some classes of problems.

- van Beek (1992) shows that path consistency induces global consistency if the constraints are *row convex.*[3]

- van Beek and Dechter (1994) show that if the constraints are $m$-tight[4], then strong $(m+2)$-consistency induces global consistency.[5]

- A tractable class of binary constraints is identified by Cooper *et al.* (1994), 0/1/all constraints. If for any pair of variables $X_i, X_j$, every value of $D_i$ is supported by either 0, 1 or *all* values of $D_j$, then there exists an algorithm, called ZOA (Zero/One/All), which either finds a solution, or informs that no solution exists in polynomial time (namely $O(ed(n+d))$). It is also shown that any other class of constraints closed under permutation and restriction of the domains is NP-complete.

- Kirousis (1993) presents a polynomial algorithm to solve CSPs with *implicational* constraints (which, for binary constraints, are the same as 0/1/all constraints[6]).

1. Which should in fact be called *bijective.*
2. It is also shown that, in the field of constraint logic programming over finite domains, for a restricted class of basic constraints, node- and arc-consistency algorithms provide a decision procedure whose complexity is the same as the complexity of AC-5, namely $O(ed)$.
3. A binary constraint $C_{ij}$ is row convex iff for any $a_i \in D_i$, all its supports are consecutive in $D_j$, provided there exists an (implicit) ordering of the domains.
4. A binary constraint $C_{ij}$ is $m$-tight iff every value $a_i \in D_i$ is supported by either at most $m$ values in $D_j$, or *all* values of $D_j$, and, conversely, every value $a_j \in D_j$ either has at most $m$ supports in $D_i$ or is supported by *all* values of $D_i$.
5. They furthermore show how this result applies to non-binary networks: for any CSP whose constraints have arity $r$ or less and are at most $m$-tight (an $r$-ary constraint is $m$-tight iff all its binary projections are $m$-tight), if the CSP is strongly $((m+1)(r-1)+1)$-consistent, then it is globally consistent.
6. For the sake of simplicity we will not present non-binary implicational constraints (for which an $O(d^2r^2ne)$ algorithm is presented), nor a parallel algorithm, also studied in his paper.

A first comment about those results: a class of binary constraints is a subclass of all those presented above, namely *bijective* constraints: they are 1-tight, 0/1(/all), and row convex. Bijective constraints are known to be tractable. Notice that constraints called *functional* by van Beek (1992) and by Van Hentenryck *et al.* (1992) are actually *bijective* constraints[7]. On the other hand, a functional constraint $x_i \rightarrow x_j$ is neither necessarily 1-tight, nor 0/1/all (see for example constraint $w = v^2$ in Figure 2), but $R_{ij}$ remains row convex (even though $R_{ji}$ may not be). But achieving path consistency in a network whose constraints are all functional may create non functional constraints (or transform some functional constraint into non functional ones). van Beek's result does not apply anymore (since there are non row convex constraints). Functional constraints, unlike bijective constraints, are intractable in general case.

The second comment is more important: all of those results assume that *all* of the constraints of the network belong to a given class (*e.g.*, $m$-tight, 0/1/all, row convex...), and only apply in this case. The difference, and, in our opinion, an interesting improvement, is that the results we present in this paper also apply when only *some* of the constraints are functional.

In a previous paper (David, 1993), it is shown that in a path consistent CSP, any consistent instantiation can be extended to a solution if, for every non-instantiated variable, there exists a sequence of functional or bijective constraints from an (any) instantiated variable to this non-instantiated variable. A subset of the variables is thus defined, such that any of its consistent instantiation can be linearly extended to a solution. Actually, path consistency is not necessary: a weaker local consistency, *pivot consistency*, is sufficient, as we now propose to show.

### 2.3 Functional Constraints and Directed Graphs

If a CSP $\mathcal{P}$ is functional, its constraint graph can be divided into two subsets. $C_f$ contains the (directed) arcs representing the functional constraints, where an arc is directed from $x_i$ to $x_k$ iff $x_i \rightarrow x_k$, and $C_o$ contains the edges (*i.e.*, undirected), for non functional constraints. We denote $\mathcal{G} = (X, C_f \cup C_o)$ this graph, and $\mathcal{G}_f = (X, C_f)$ its directed subgraph.

Before going further, we need to recall some graph theory notions: the *descendant* of a vertex (variable), and the *root* of a directed graph:

**Definition 2.3 (Berge, 1970)** *A vertex $x_k$ is a* descendant *of $x_i$ in a directed graph $(X, C_f)$ iff there exists in $C_f$ a path $(x_i, x_{j_1}, \ldots, x_{j_q}, \ldots, x_{j_p}, x_k)$ such that $x_i \rightarrow x_{j_1}, \ldots, x_{j_q} \rightarrow x_{j_{q+1}}, \ldots, x_{j_p} \rightarrow x_k$, $\forall q \in 1..p-1$.*

*A vertex $x_r$ is a* root *of a directed graph $(X, C_f)$ iff any other vertex is a descendant from $x_r$.*

By extension, we define a *root set* of a directed graph. It is easy to find out that a *root* is a root set which contains a single vertex.

**Definition 2.4** *A subset $\mathcal{R}$ of $X$ is a* root set *of a directed graph $(X, C_f)$ iff any vertex of $X - \mathcal{R}$ is a descendant from an element of $\mathcal{R}$. $\mathcal{R}$ is called* minimal *iff there does not exist a root set $\mathcal{R}'$ such that $\mathcal{R}' \subset \mathcal{R}$. $\mathcal{R}$ is called* minimum *iff there does not exist a root set $\mathcal{R}'$ such that $|\mathcal{R}'| < |\mathcal{R}|$.*

7. A constraint $c_{ij}$ is bijective iff it is functional from $x_i$ to $x_j$ ($x_i \rightarrow x_j$) *and* from $x_j$ to $x_i$ ($x_j \rightarrow x_j$).

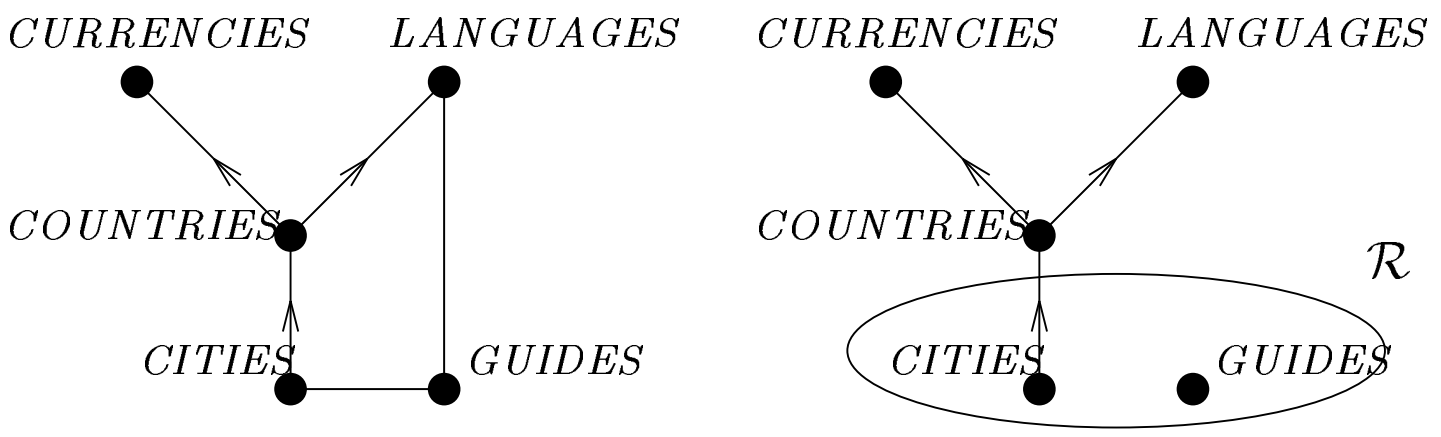

$\mathcal{G} = (X, C_f \cup C_o)$ $\qquad$ $\mathcal{G}_f = (X, C_f)$

Figure 3: Subgraph and root set of $\mathcal{P}_{ex}$

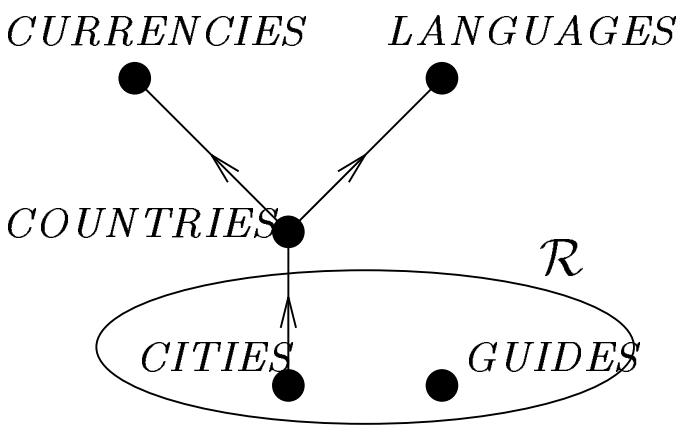

*GUIDES, CITIES, COUNTRIES, CURRENCIES, LANGUAGES* or *CITIES, GUIDES, COUNTRIES, LANGUAGES, CURRENCIES* are both $\mathcal{R}$-compatible.

*GUIDES, CITIES, LANGUAGES, COUNTRIES, CURRENCIES* is prefixed by $\mathcal{R}$, but it is not $\mathcal{R}$-compatible: *LANGUAGES* has no ancestor $x_i$ in this ordering such that $x_i \rightarrow$ *LANGUAGES*.

*GUIDES, LANGUAGES, COUNTRIES, CURRENCIES, CITIES* is not $\mathcal{R}$-compatible: it is not prefixed by $\mathcal{R}$.

Figure 4: $\mathcal{R}$-compatible and non $\mathcal{R}$-compatible orderings of $\mathcal{P}_{ex}$

Figure 3 illustrates the notions of directed subgraph and of root set (here, a *minimum* root set) applied to our example $\mathcal{P}_{ex}$. Notice that the arc is only directed from *COUNTRIES* to *CURRENCIES*, and not from *CURRENCIES* to *COUNTRIES* though we have both *COUNTRIES* $\rightarrow$ *CURRENCIES* and *CURRENCIES* $\rightarrow$ *COUNTRIES*. This is only for the sake of simplicity: this information will not be useful in the following.

With a root set $\mathcal{R}$, we associate $\mathcal{R}$*-compatibility*, a property that can be verified by an ordering of $X$:

**Definition 2.5** *An ordering of* $X$ $x_1, x_2, \ldots, x_n$ *is called* $\mathcal{R}$-compatible *iff:*

1. $\forall i \leq |\mathcal{R}|$, $x_i \in \mathcal{R}$, *and*
2. $\forall k > |\mathcal{R}|$, $\exists j < k$ *such that* $x_j \rightarrow x_k$.

In other words, the first variables in this ordering are those which belong to $\mathcal{R}$ (we then say that this ordering is *prefixed* by $\mathcal{R}$), and any other variable $x_k$ has at least one ancestor $x_j$ in the ordering ($x_j$ is ordered before $x_k$) such that $x_j \rightarrow x_k$ (see Figure 4).

## 3. Pivot Consistency

First of all, why did we call it *pivot consistency*? Pivot consistency depends on the assignment ordering[8]. On every step of the assignment, two elements have to be distinguished: firstly, the set of the formerly instantiated variables (the *instantiated set*), and secondly the

8. Similarly to directional path consistency and adaptative consistency (Dechter & Pearl, 1988).

next variable to be instantiated. One can consider pivot consistency as a property between each instantiated set and the next variable to be instantiated. For each of these sets, a constraint has a particular role to play: the consistency checks turn on it. This constraint is thus called the *pivot* of the set.

### 3.1 Presentation

Further in this section, we will introduce the definition of pivot consistency in 3 steps: consistency between 3 variables, then pivot of a subset of $X$, and finally pivot consistency of the CSP.

**Definition 3.1** *Let $x_i, x_j, x_k \in X$ such that $c_{ik}, c_{jk} \in C$. $c_{ik}$ and $c_{jk}$ are $x_k$-*compatible *iff any tuple of $R_{ij}$ has at least one support in $D_k$ for $c_{ik}$ and $c_{jk}$. Formally, $\forall (a_i, a_j) \in R_{ij}$, $\exists a_k \in D_k$ s.t. $(a_i, a_k) \in R_{ik}$ and $(a_j, a_k) \in R_{jk}$*

Four comments arise from this definition:

1. The existence of a constraint between $x_i$ and $x_j$ is not compulsory: the relation $R_{ij}$ may be universal.
2. $c_{ik}$ and $c_{jk}$ are not necessarily different. If not, $c_{ik}$ must be $x_k$-*compatible* with itself; in that case we assimilate the pair $(a_i, a_i)$ of the relation $R_{ii}$ to the value $a_i$ of the domain $D_i$.
3. This definition may be seen as a local version of strong 3-consistency (2 and 3 consistencies):
   - 2-consistency: it is due to the remark above: any value in $D_i$ must have a support in $D_k$.
   - 3-consistency: any consistent instantiation of $\{x_i, x_j\}$ may be extended to a third variable, here $x_k$.
4. Knowing that path consistency is equivalent to 3-consistency, we can deduce that path consistency can be rewritten in terms of $x_k$-compatibility:

   *A CSP is path consistent iff for all $x_i, x_j, x_k$ pairwise distinct, $\{x_i, x_k\}$ and $\{x_j, x_k\}$ are both $x_k$-compatible.*

We already said that some constraints called the *pivots* have a particular part to play; we now define them:

**Definition 3.2** *Let $Y \subset X$ and $c_{ik} \in C$ s.t. $x_i \rightarrow x_k$, $x_k \in X - Y$ and $x_i \in Y$. $x_i \rightarrow x_k$ is a* pivot *of $Y$ iff $\forall x_j \in Y$ s.t. $c_{jk} \in C$, $c_{ik}$ and $c_{jk}$ are $x_k$-compatible.*

In other words, given any proper subset $Y$ of $X$, a functional constraint $x_i \rightarrow x_k$ "coming out" from $Y$ ($x_i \in Y$ and $x_k \in X - Y$) is a pivot of $Y$ if and only if for any consistent instantiation $(a_i, a_j)$ of $x_i$ and of any other variable $x_j$ in $Y$, there exists at least one value $a_k$ in $D_k$ such that $(a_i, a_j)$ can be extended to a consistent instantiation $(a_i, a_j, a_k)$ of $\{x_i, x_j, x_k\}$. As for functional constraints, we call $x_i$ the *origin* and $x_k$ the *target* of the pivot $x_i \rightarrow x_k$.

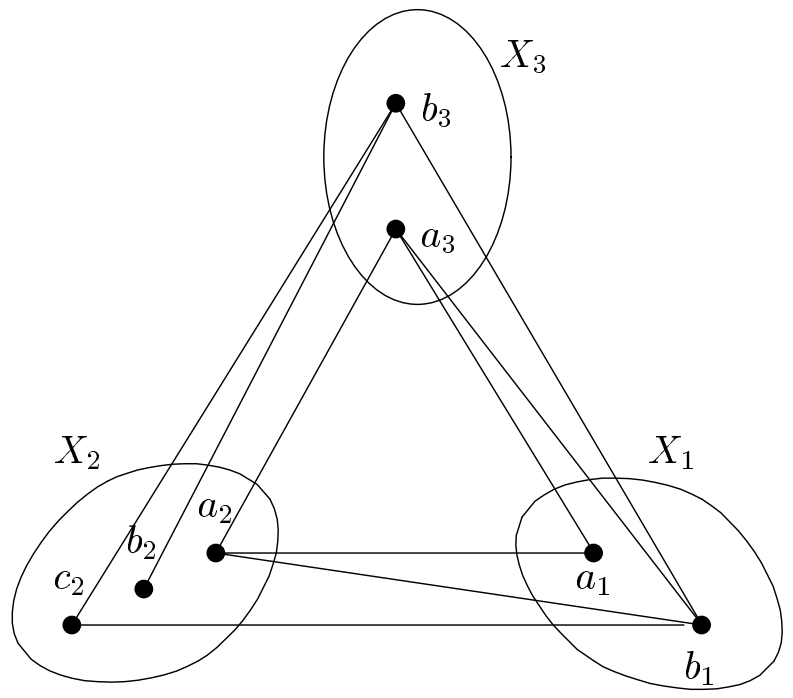

$\mathcal{R} = \{x_1, x_2\}$
The pivot of $\{x_1, x_2\}$ is $x_2 \to x_3$

This CSP is pivot consistent: the 3 pairs of $R_{12}$, that is $(a_1, a_2)$, $(b_1, a_2)$ and $(b_1, c_2)$ all have a support in $D_3$ (respectively $a_3$, $a_3$ and $b_3$).

On the other hand, it is neither path consistent, since $(b_2, b_3)$ has no support in $D_1$, nor arc consistent, since $b_2$ has no support in $D_1$.

Figure 5: A pivot consistent CSP

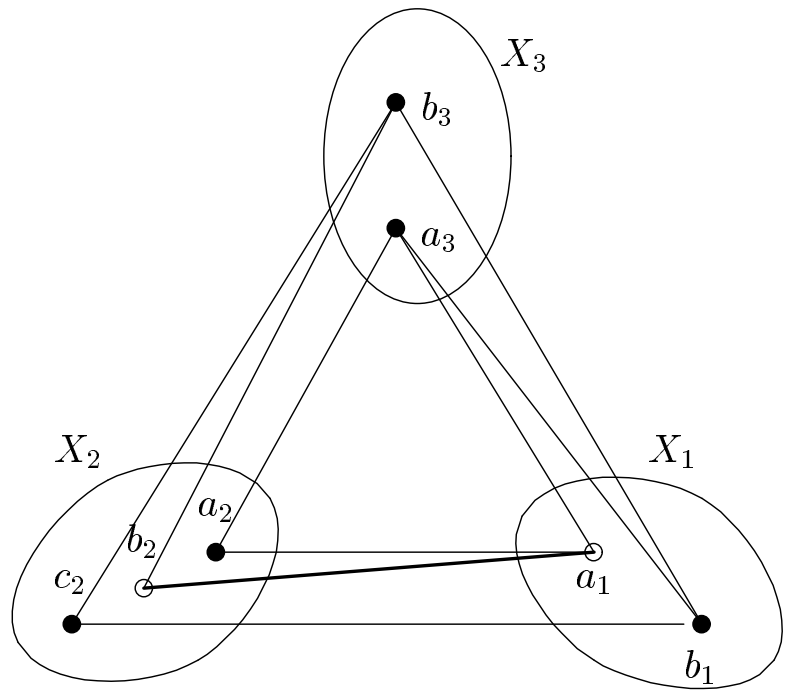

$\mathcal{R} = \{x_1, x_2\}$
The pivot candidate of $\{x_1, x_2\}$ is $x_2 \to x_3$.

This CSP is not pivot consistent: $(a_1, b_2)$ has no support in $D_3$.

Figure 6: A non pivot consistent CSP

Now that the pivot of a subset $Y$ of $X$ has been defined, we can introduce the notion of *pivot consistency* of a CSP:

**Definition 3.3** *Given a* CSP $\mathcal{P} = (X, D, C, R)$, *a root set* $\mathcal{R}$ *of* $(X, C)$ *and an* $\mathcal{R}$*-compatible assignment ordering* $x_1, x_2, \ldots, x_n$, *we say* $\mathcal{P}$ *is* pivot consistent *w.r.t. this ordering iff* $\forall k > r = |\mathcal{R}|$, $\exists h < k$ *s.t.* $x_h \to x_k$ *is a pivot of* $Y_{k-1} = x_1, x_2, \ldots, x_{k-1}$.

Informally, every variable which does not belong to $\mathcal{R}$ is the target of a pivot whose origin is before it in the assignment ordering. Figures 5 and 6 show a pivot consistent CSP and a non pivot consistent CSP.

Pivot consistency of a CSP relies on the existence of a set of functional constraints possessing particular properties: the pivots (Definition 3.3 above). Thus, a *minimum* set of functional constraints can be characterized, so that the CSP is pivot consistent. This is the purpose of the three conditions stated in Definition 3.4:

- condition 1 ensures there exists a pivot for each $Y_{k-1}$, $k > r$: the network is therefore pivot consistent;

- conditions 2 and 3 leave unnecessary constraints out: only one pivot needs to "target" each $x_k \in X - \mathcal{R}$, and no pivot is required inside $\mathcal{R}$.

**Definition 3.4** *Given a* CSP $\mathcal{P} = (X, D, C = C_f \cup C_o, R)$*, a root set* $\mathcal{R}$*, an* $\mathcal{R}$*-compatible assignment ordering* $x_1, x_2, \ldots, x_n$*, and a set of functional constraints* $P \subseteq C_f$*, if the following three conditions are satisfied*

1. $\forall x_k \in X - \mathcal{R}$, $\exists x_h \to x_k \in P$ *s.t.* $h < k$ *and* $x_h \to x_k$ *is pivot of* $Y_{k-1}$
   *(any variable of* $X - \mathcal{R}$ *is the target of a pivot)*
2. $\forall x_k \in X - \mathcal{R}$, $\{x_h \to x_k \in P$ *and* $x_j \to x_k \in P\} \Rightarrow h = j$
   *(any variable of* $X - \mathcal{R}$ *is the target of at most one pivot)*
3. $\forall x_j \in \mathcal{R}$, $\nexists x_i \to x_j \in P$
   *(no variable of* $\mathcal{R}$ *is the target of a pivot)*

*then* $P$ *is called a* pivot set *of the* CSP*, and* $\mathcal{P}$ *is pivot consistent.*

We have introduced a new local consistency, pivot consistency. However, as for any local consistency, a CSP generally does not satisfy it. The problem has then to be *filtered* in order to obtain its *pivot consistent closure*, which is presented in the next section. This filtering is achieved w.r.t. a given pivot set $P$: this is why we now propose the definition of pivot consistency *with respect to a pivot set* $P$:

**Definition 3.5** *Given a* CSP $\mathcal{P} = (X, D, C, R)$*, a root set* $\mathcal{R}$ *of* $(X, C)$*, an* $\mathcal{R}$*-compatible assignment ordering* $x_1, x_2, \ldots, x_n$ *and a constraint set* $P \subseteq C$*, we say* $\mathcal{P}$ *is* pivot consistent with respect to $P$ and this ordering *iff* $P$ *is a pivot set of* $\mathcal{P}$*.*

### 3.2 Pivot Consistent Closure

Let us assume a CSP $\mathcal{P}$ is not pivot consistent, and that we wish to make it so. We then obtain a new problem, say $\mathcal{P}^p$, but we want this problem to meet some properties. The purpose of this section is to specify these properties. We first define the *pivot consistent closure* of a given CSP:

**Definition 3.6** $\mathcal{P}^p = (X, D^p, C^p, R^p)$ *is called the* pivot consistent closure *of the* CSP $\mathcal{P} = (X, D, C, R)$ *iff*

1. $\mathcal{P}^p \trianglelefteq \mathcal{P}$ (i.e., $\forall i \in 1..n, D_i^p \subseteq D_i$, $C \subseteq C^p$, *and* $\forall i, j \in 1..n, R_{ij}^p \subseteq R_{ij}$ *or* $c_{ij} \notin C$)*, and*
2. $\mathcal{P}^p$ *is pivot consistent, and*
3. $\mathcal{P}^p$ *is* maximal*: there does not exist* $\mathcal{P}'$ *pivot consistent such that* $\mathcal{P}^p \trianglelefteq \mathcal{P}' \trianglelefteq \mathcal{P}$ *and* $\mathcal{P}^p \neq \mathcal{P}'$*.*

Before presenting two properties of the pivot consistent closure $\mathcal{P}^p$ of a CSP $\mathcal{P}$, we introduce the following lemma, which we will use later in one of the proofs.

**Lemma 3.1** *Let* $\mathcal{P} = (X, D, C, R)$ *be a* CSP*. Let two* CSP*s* $\mathcal{P}^1 = (X, D^1, C^1, R^1) \trianglelefteq \mathcal{P}$ *and* $\mathcal{P}^2 = (X, D^2, C^2, R^2) \trianglelefteq \mathcal{P}$ *be pivot consistent w.r.t. a root set* $\mathcal{R}$*, a pivot set* $P = \{x_{Origin(k)} \to x_k, \forall k > |\mathcal{R}|\}$ *and an* $\mathcal{R}$*-compatible assignment ordering. Let the* CSP $\mathcal{P}^3 = (X, D^3 = D^1 \sqcup D^2, C^3 = C^1 \cap C^2, R^3 = R^1 \sqcup R^2)$*, with* $D^3 = D^1 \sqcup D^2 = \{D_i^3 = D_i^1 \cup D_i^2\ \forall i \in 1..n\}$ *and* $R^3 = R^1 \sqcup R^2 = \{R_{ij}^3 = R_{ij}^1 \cup R_{ij}^2\ \forall i, j \in 1..n$ *s.t.* $c_{ij}^3 \in C^3\}$*. Then firstly* $\mathcal{P}^3$ *is pivot consistent w.r.t. these same sets* $\mathcal{R}$ *and* $P$ *and this same* $\mathcal{R}$*-compatible assignment ordering, and secondly* $\mathcal{P}^1, \mathcal{P}^2 \trianglelefteq \mathcal{P}^3 \trianglelefteq \mathcal{P}$*.*

**Proof:** We first show that $\mathcal{P}^3$ is pivot consistent. Let us denote $x_{Origin(k)} \to x_k$ (or, more briefly, $x_{O(k) \to x_k}$) the pivot whose target is $x_k$ in $\mathcal{P}^1$ and $\mathcal{P}^2$, for all $k > |\mathcal{R}|$. Let us show that for all $k > |\mathcal{R}|$, $x_{O(k) \to x_k}$ is the pivot of $Y_{k-1}$ in $\mathcal{P}^3$. Let $x_j$ in $Y_{k-1}$ such that $c_{jk} \in C^1 \cap C^2$. For all $(a_j, a_{O(k)}) \in R^3_{jO(k)}$, either $(a_j, a_{O(k)}) \in R^1_{jO(k)}$ or $(a_j, a_{O(k)}) \in R^2_{jO(k)}$. If $(a_j, a_{O(k)}) \in R^1_{jO(k)}$: as $\mathcal{P}^1$ is pivot consistent, there exists $a_k$ in $D^1_k$ such that $(a_j, a_k) \in R^1_{jk}$ (therefore $R^3_{jk}$) and $(a_{O(k)}, a_k) \in R^1_{O(k)k}$ (therefore $R^3_{O(k)k}$). It is the same if $(a_j, a_{O(k)}) \in R^2_{jO(k)}$. $c_{jk}$ and $x_{O(k)} \to x_k$ are consequently $x_k$-compatible: $x_{O(k)} \to x_k$ is pivot of $Y_{k-1}$ in $\mathcal{P}^3$.
For all $k > |\mathcal{R}|$, there exists $x_{O(k)} \to x_k$ pivot of $Y_{k-1}$: $\mathcal{P}^3$ is pivot consistent.

Let us now show that $\mathcal{P}^1, \mathcal{P}^2 \trianglelefteq \mathcal{P}^3 \trianglelefteq \mathcal{P}$:

- $\forall i \in 1..n$, $D^3_i = D^1_i \cup D^2_i$, so $D^1_i \subseteq D^3_i$ and $D^2_i \subseteq D^3_i$. (1)

  Furthermore, we have $D^1_i \subseteq D_i$ and $D^2_i \subseteq D_i$; consequently, $D^3_i \subseteq D_i$. (1′)

- $C^3 = C^1 \cap C^2$, so $C^3 \subseteq C^1$ and $C^3 \subseteq C^2$. (2)

  Also, since $C \subseteq C^1$ and $C \subseteq C^2$, we have $C \subseteq C^3 = C^1 \cap C^2$. (2′)

- $\forall i, j \in 1..n$ such that $c^3_{ij} \in C^3$, $R^3_{ij} = R^1_{ij} \cup R^2_{ij}$ so $R^1_{ij} \subseteq R^3_{ij}$ and $R^2_{ij} \subseteq R^3_{ij}$. (3)

- $\forall i, j \in 1..n$ such that $c_{ij} \in C$, do we have $R^3_{ij} \subseteq R_{ij}$? Let $c_{ij} \in C$. Then $c^3_{ij} \in C^3$ ($C \subseteq C^3$). By construction of $R^3$, we therefore have $R^3_{ij} = R^1_{ij} \cup R^2_{ij}$. Now $R^1_{ij} \subseteq R_{ij}$ and $R^2_{ij} \subseteq R_{ij}$, which implies $R^3_{ij} \subseteq R_{ij}$. (3′)

We deduce from (1), (2) and (3) that $\mathcal{P}^1 \trianglelefteq \mathcal{P}^3$ and $\mathcal{P}^2 \trianglelefteq \mathcal{P}^3$, and $\mathcal{P}^3 \trianglelefteq \mathcal{P}$ from (1′), (2′) and (3′). □

We can now present the properties we announced in the beginning of this section.

**Property 3.2** *There only exists one maximal $\mathcal{P}^p \trianglelefteq \mathcal{P}$ pivot consistent, for a given root set $\mathcal{R}$, a given pivot set and a given assignment ordering.*

**Proof:** Suppose that $\mathcal{P}^p$ is not unique: thus, there exists $\mathcal{P}' = (X, D', C', R') \trianglelefteq \mathcal{P}$ such that $\mathcal{P}'$ is maximal and pivot consistent, and $\mathcal{P}' \neq \mathcal{P}^p$. Let us build the CSP $\mathcal{P}'' = (X, D^p \sqcup D', C^p \cap C', R^p \sqcup R')$. This CSP is pivot consistent and $\mathcal{P}^p \trianglelefteq \mathcal{P}'' \trianglelefteq \mathcal{P}$ (Lemma 3.1): $\mathcal{P}^p$ is not maximal, which contradicts the previous assumption. □

The second property of $\mathcal{P}^p$ is perhaps the most important:

**Property 3.3** *Both* CSPs *$\mathcal{P}$ and $\mathcal{P}^p$ are equivalent, in that they have the same set of solutions.*

**Proof:** Let us denote $\mathcal{S}(\mathcal{P})$ and $\mathcal{S}(\mathcal{P}^p)$ their respective sets of solutions. Obviously, $\mathcal{S}(\mathcal{P}^p) \subseteq \mathcal{S}(\mathcal{P})$. Let us now show that $\mathcal{S}(\mathcal{P}) \subseteq \mathcal{S}(\mathcal{P}^p)$: suppose there exists $I_n = (s_1, \ldots, s_n)$ solution of $\mathcal{P}$ which is not a solution of $\mathcal{P}^p$. Then $\exists i, j \in 1..n$ such that $(s_i, s_j) \notin R^p_{ij}$. Now $(s_i, s_j) \in R_{ij}$ ($I_n$ is a solution of $\mathcal{P}$): as $\mathcal{P}^p$ is maximal, that means that $(s_i, s_j)$ has no support for at least one constraint: $\exists k$ s.t. $\forall a_k \in D_k$, $(s_i, a_k) \notin R_{ik}$ or $(s_j, a_k) \notin R_{jk}$. We therefore have $(s_i, s_k) \notin R_{ik}$ or $(s_j, s_k) \notin R_{jk}$: $I_n$ is not a solution of $\mathcal{P}$, that is $\mathcal{S}(P) \subseteq \mathcal{S}(\mathcal{P}^p)$.
□

We now know that there exists a unique CSP $\mathcal{P}'$, called the *pivot consistent closure* of $\mathcal{P}$, such that $\mathcal{P}' \trianglelefteq \mathcal{P}$, $\mathcal{P}'$ is maximal and $\mathcal{P}'$ is pivot consistent. Moreover, this CSP has the same set of solutions as $\mathcal{P}$, the problem it comes from. But, even though it is interesting to know that such a CSP exists, it seems reasonable to wish to obtain it: that is the purpose of the next section, which proposes an algorithm achieving a pivot consistent filtering.

### 3.3 A Filtering Algorithm

First, a notation: the set of the functional constraints $x_h \to x_k$ chosen to be the pivots of each $Y_k$ is called a *set of pivot candidates*, and denoted $PC$. After the filtering, $PC = P$, the pivot set. We will suppose in the course of this section that both the assignment ordering and $PC$ are known; we explain how they are obtained in appendix A.2.

The algorithm is composed of several procedures. Their different levels represent the three steps followed to define pivot consistency. We present these procedures in an ascending way:

1. Make the constraints $c_{jk}$ and $c_{hk} = x_h \to x_k$ $x_k$-compatible.
2. Compute a pivot $x_h \to x_k$ for all $x_k$ in $X - \mathcal{R}$
3. Achieve pivot consistency for the CSP.

Procedure `Compatible` $(x_h, x_k, x_j)$ makes constraints $x_h \to x_k$ and $c_{jk}$ $x_k$-compatible: it removes from $R_{hj}$ those tuples which do not have a common support in $D_k$ for these two constraints. If necessary, it creates the constraint $c_{hj}$ between the variables $x_h$ and $x_j$.

```
Procedure Compatible(x_h, x_k, x_j)
  begin
    for all (a_h, a_j) ∈ R_hj do
      a_k ← f_hk(a_h)^9
      if (a_h, a_k) ∉ R_hk or (a_j, a_k) ∉ R_jk then Suppress (a_h, a_j) from R_hj
    end for
  end
```

Procedure `Pivot` $(x_h, x_k)$ makes all constraints $c_{jk}$ containing $x_k$ and such that $x_j \in Y_{k-1}$ $x_k$-compatible with $x_h \to x_k$ by successive calls to the subroutine `Compatible` $(x_h, x_k, x_j)$. After its computation, $x_h \to x_k$ is therefore a pivot of $Y_{k-1}$.

```
Procedure Pivot(x_h, x_k)
  begin
    for all x_j ∈ Y_{k-1} s.t. c_jk ∈ C do Compatible(x_h, x_k, x_j)
  end
```

The CSP is pivot consistent iff any variable $x_k$ of $X - \mathcal{R}$ is the target of a pivot of $Y_{k-1}$.

9. By convention, if $a_h$ has no support in $D_k$ for $c_{hk}$, we denote $f_{hk}(a_h) = \varepsilon$, and, of course $(a_h, \varepsilon) \notin R_{hk}$ and $(a_j, \varepsilon) \notin R_{jk}$.

```
Pivot consistent filtering
      begin
        for k ← n to r+1 do
          x_h ← Origin of the pivot candidate whose target is x_k
          Pivot (x_h, x_k)
        end for
      end
```

**Proposition** *This algorithm achieves pivot consistency (the resulting problem is the pivot consistent closure of $\mathcal{P}$ w.r.t. the chosen data).*

**Proof:** We clearly see that after computing `Pivot` $(x_h, x_k)$, $x_h \rightarrow x_k$ is a pivot of $Y_{k-1}$. Moreover, only the necessary suppressions are performed: according to the definition of `Compatible`, a pair is suppressed from a relation only if it has no support in the target of the current pivot candidate.

We consequently simply have to show that any suppression due to `Pivot` $(x_h, x_k)$ has no influence on the pivots $x_g \rightarrow x_l$ for $l > k$. More precisely, if $x_g \rightarrow x_l$ is a pivot of $Y_{l-1}$ before computing `Pivot` $(x_h, x_k)$, then it will remain so.

For the sake of brevity, we will denote in the course of this proof $Pv(k)$ the computation of the procedure `Pivot` $(x_h, x_k)$. Let $l$ be such that $x_g \rightarrow x_l$ is a pivot of $Y_{l-1}$ before $Pv(k)$. Suppose $x_g \rightarrow x_l$ is no longer a pivot of $Y_{l-1}$ after $Pv(k)$. Consequently, there exists $j < l$ such that $c_{jl}$ and $c_{gl}$ are not $x_l$-compatible: $\exists (a_j, a_g) \in R_{jg}$ s.t. $(a_j, a_l) \notin R_{jl}$ or $(a_g, a_l) \notin R_{gl}$. $Pv(k)$ has thus suppressed $(a_j, a_l)$ from $R_{jl}$ or $(a_g, a_l)$ from $R_{gl}$. This is impossible, since by definition procedure `Pivot` $(x_h, x_k)$ only modifies constraints $c_{hi}$, with $h, i < k$, so *a fortiori* $h, i < l$ $(k < l)$. So, $Pv(k)$ does not influence pivots $x_g \rightarrow x_l$ for $l > k$. This algorithm therefore computes the pivot consistent closure of a CSP. □

Notice that unlike arc or path consistencies which only need knowledge about the CSP, the pivot consistent closure is in addition defined w.r.t. a root set, a pivot set and an assignment ordering: Figure 7 shows how the choice of the pivot(s) may influence the filtered problem.

### 3.4 Application to the Example

In this section, we illustrate the filtering algorithm using the travel agency example. We first recall the initial problem and expose the data necessary to the algorithm (root set $\mathcal{R}$, set of pivot candidates $PC$ and $\mathcal{R}$-compatible ordering).

$\mathcal{R} = \{$*GUIDES, CITIES*$\}$

$PC = \{$*CITIES* → *COUNTRIES*,
*COUNTRIES* → *CURRENCIES*,
*COUNTRIES* → *LANGUAGES*$\}$

An $\mathcal{R}$-compatible ordering is:
*GUIDES, CITIES, COUNTRIES, CURRENCIES, LANGUAGES*

In order to achieve pivot consistency, we need to compute `Pivot` (*COUNTRIES,LANGUAGES*), `Pivot` (*COUNTRIES,CURRENCIES*) and `Pivot` (*CITIES,COUNTRIES*).

$\mathcal{R} = \{x_1, x_2, x_3\}$
The three possible pivots are $x_1 \to x_4$, $x_2 \to x_4$ and $x_3 \to x_4$.
The assignment ordering we have chosen is $x_1, x_2, x_3, x_4$.

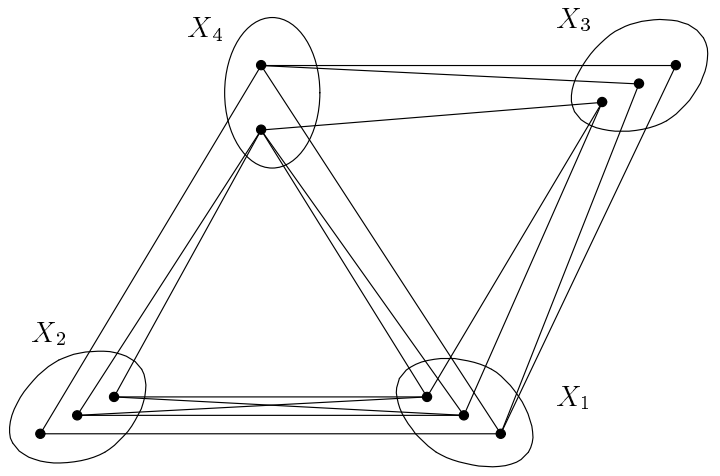

If we choose the pivot $x_1 \to x_4$ the filtering does not suppress anything; the resulting problem $\mathcal{P}_1$ is therefore the same as the initial one $\mathcal{P}$.

If we choose the pivot $x_2 \to x_4$ the filtering creates the constraint $c_{23}$, and does not change anything else; we obtain the problem $\mathcal{P}_2$:

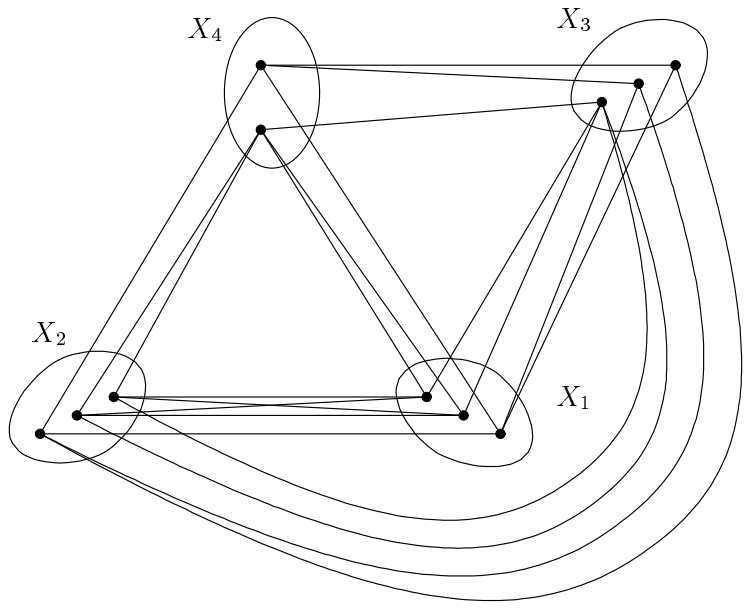

Figure 7: Pivot consistency: influence of the pivot

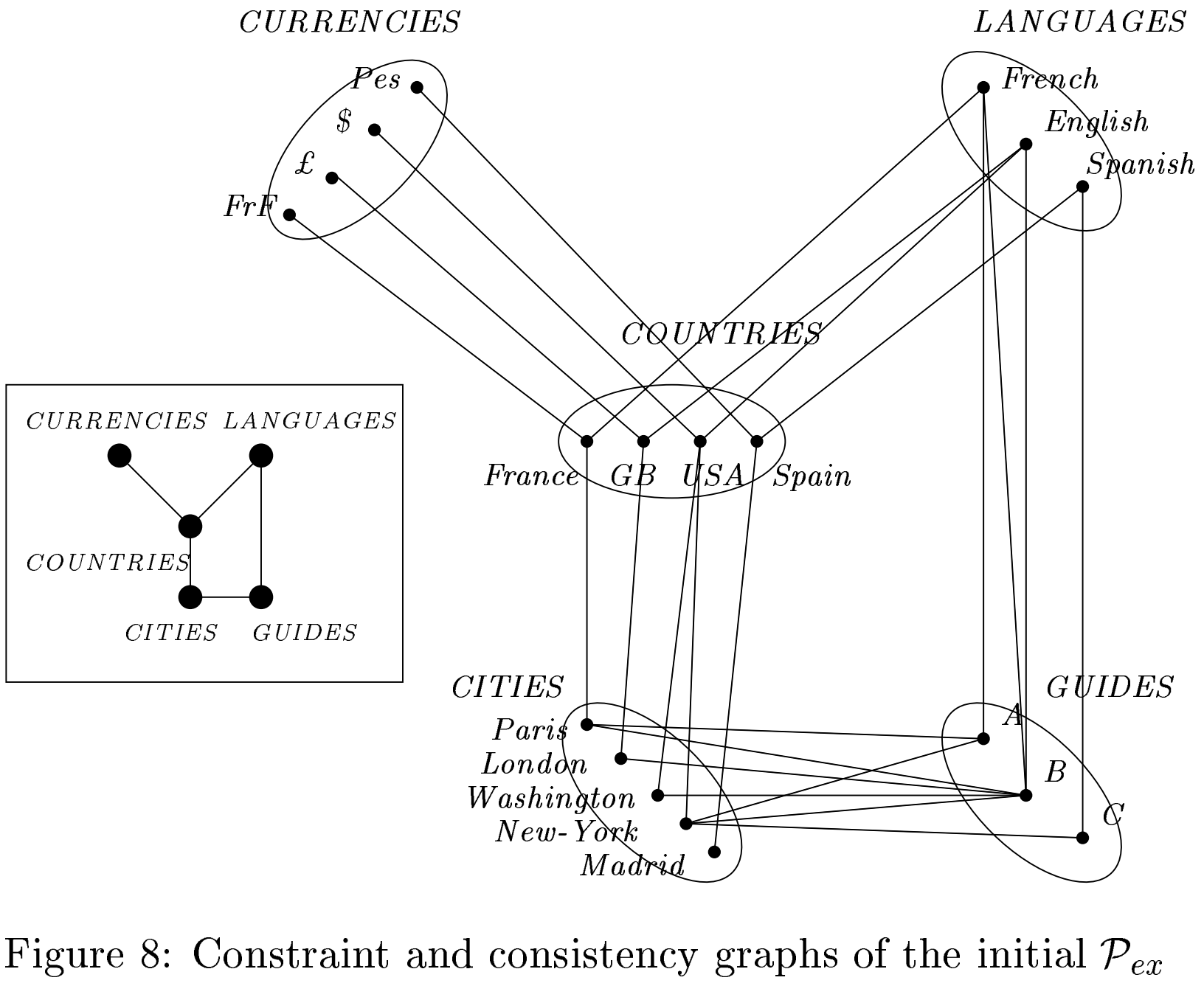

Figure 8: Constraint and consistency graphs of the initial $\mathcal{P}_{ex}$

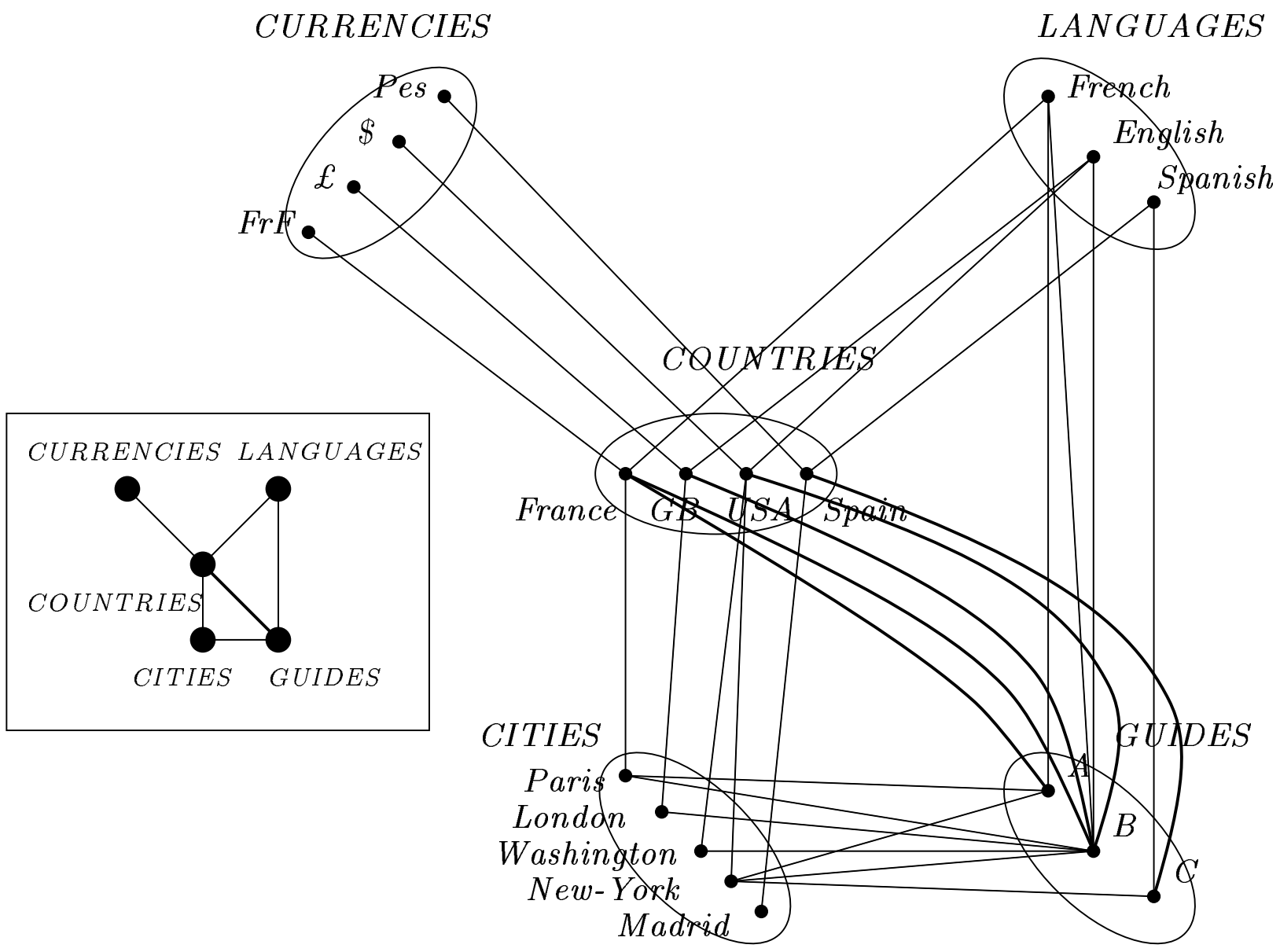

Figure 9: After `Pivot` (*COUNTRIES,LANGUAGES*)

We now detail these computations:

- Computation of `Pivot` (*COUNTRIES,LANGUAGES*) (Figure 9):
  We perform `Compatible` (*COUNTRIES,LANGUAGES,COUNTRIES*) which does not modify anything, and `Compatible` (*COUNTRIES,LANGUAGES,GUIDES*), which creates the constraint {*GUIDES, COUNTRIES*}.

- Computation of `Pivot` (*COUNTRIES,CURRENCIES*):
  We perform `Compatible` (*COUNTRIES,CURRENCIES,COUNTRIES*) which does not modify anything.

- Computation of `Pivot` (*CITIES,COUNTRIES*) (Figure 10):
  We perform `Compatible` (*CITIES,COUNTRIES,CITIES*) which does not change anything, and `Compatible` (*CITIES,COUNTRIES,GUIDES*), which modifies the constraint {*GUIDES, CITIES*}.

### 3.5 Pivot Consistency *vs* Path Consistency

We previously said that arc-and-path consistency was not necessary to give some properties to a functional CSP, and that pivot consistency was sufficient. We first show in this section that pivot consistency is a weakened form of path consistency, and in section 4 we will present some properties and a method for solving functional CSPs based upon them.

Note that, since pivot consistency is directed, one could wish to compare it not only with "full" path consistency, but also with *directional* path consistency. We therefore propose to present relationships between pivot consistency and both path consistency and directional path consistency. We will thus see that pivot consistency is a restricted version of directional path consistency as well.

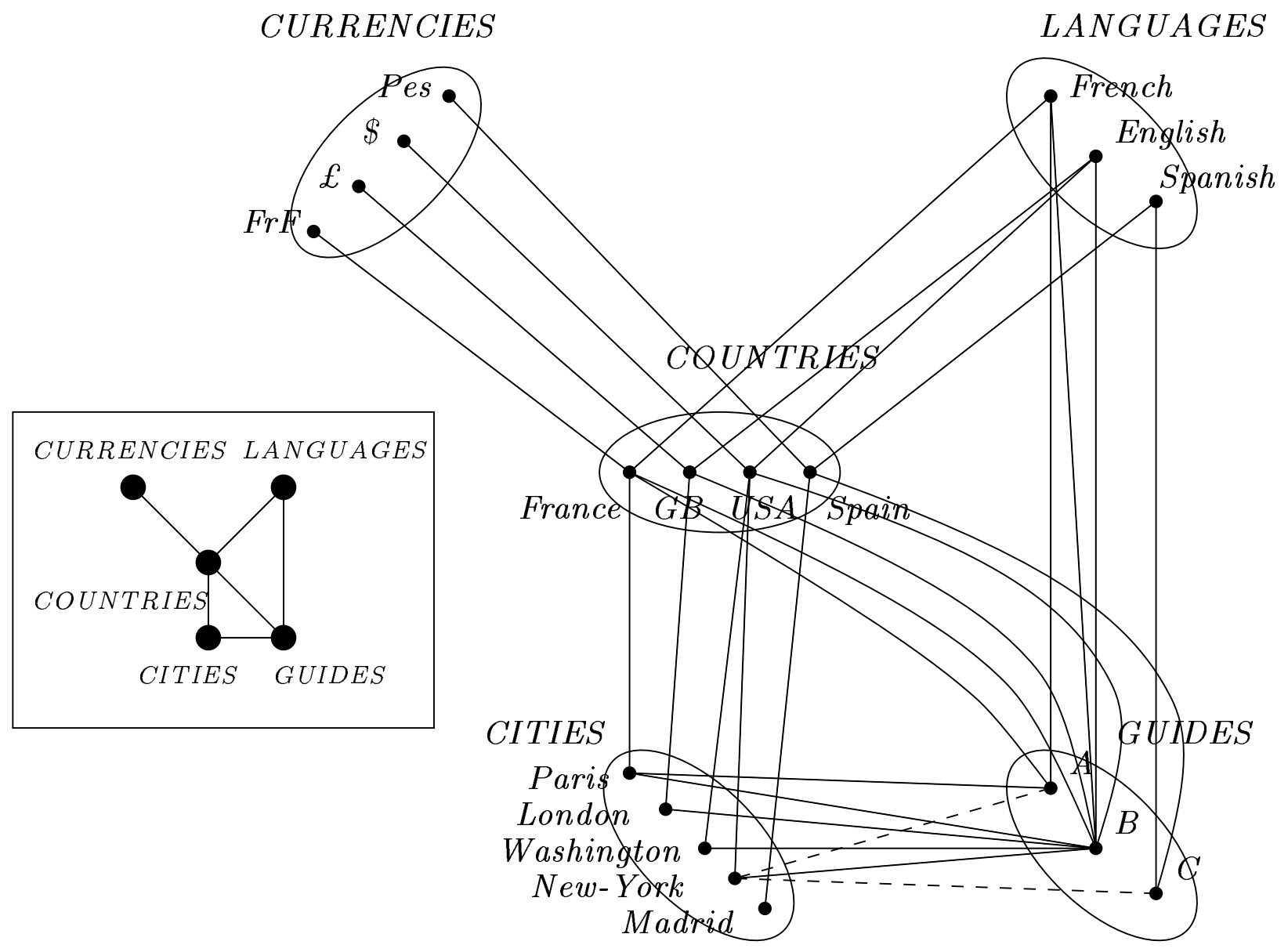

Figure 10: After `Pivot` (*CITIES,COUNTRIES*): end of the filtering

First, let us recall a remark we made when defining $x_k$-compatibility:

> *A* CSP *is path consistent iff for all* $x_k$ *in* $X$ *and for all* $x_h, x_i$ *in* $X$*,* $\{x_h, x_k\}$ *and* $\{x_i, x_k\}$ *are* $x_k$*-compatible.*

A slight difference appears with directional path consistency:

> *A* CSP *is directional path consistent iff for all* $x_k$ *in* $X$ *and for all* $x_h, x_i$ *in* $X$ such that $h, i < k$*,* $\{x_h, x_k\}$ *and* $\{x_i, x_k\}$ *are* $x_k$*-compatible.*

Pivot consistency does not need so many requirements. First, $x_k$-compatibility is only required for variables in $X - \mathcal{R}$. Furthermore, for each of these $x_k$, not all of the possible pairs of constraints are filtered: as this consistency is directed, $x_k$-compatibility is only achieved for constraints whose first variable precedes $x_k$ (the second one) in the assignment ordering. Finally, the filtering is only achieved w.r.t. the pivot candidate whose target is $x_k$ (say $x_{h_k} \to x_k$). In other words,

> *A* CSP *is pivot consistent iff for all* $x_k$ *in* $X - \mathcal{R}$ *and for all* $x_i$ *in* $X$ *such that* $i < k$*,* $\{x_{h_k}, x_k\}$ *and* $\{x_i, x_k\}$ *are* $x_k$*-compatible, provided that* $x_{h_k} \to x_k \in PC$*.*

To sum up, one can roughly say that for each $x_k$, path consistency needs to consider $O(n^2)$ triangles, directional path consistency $O(k^2)$ and pivot consistency $O(k)$. The difference between path and pivot consistencies clearly appears on the travel agency example: only two constraints are altered. The first one ({*GUIDES, COUNTRIES*}) is created, and the second one ({*GUIDES, CITIES*}) is modified, whereas path consistency modified the five existing constraints, and created the five other possible constraints, making the constraint graph complete (see Figure 11). However, this example does not highlight any difference between pivot consistency and directional path consistency. Figure 12 presents how pivot consistency and directional path consistency may differently modify the constraint graph of a CSP when processing a node $x_k$.

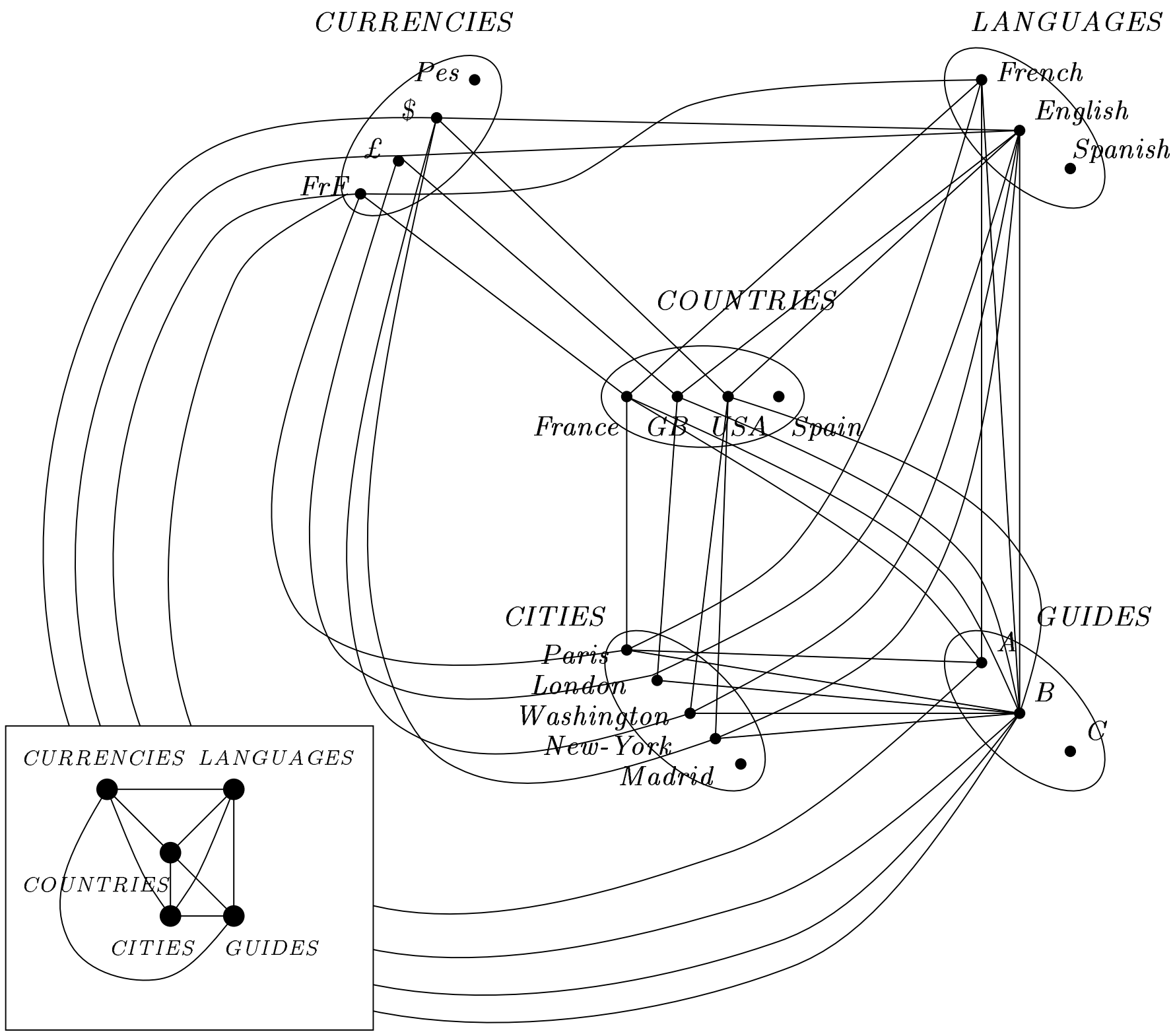

Figure 11: Path consistent closure of $\mathcal{P}_{ex}$

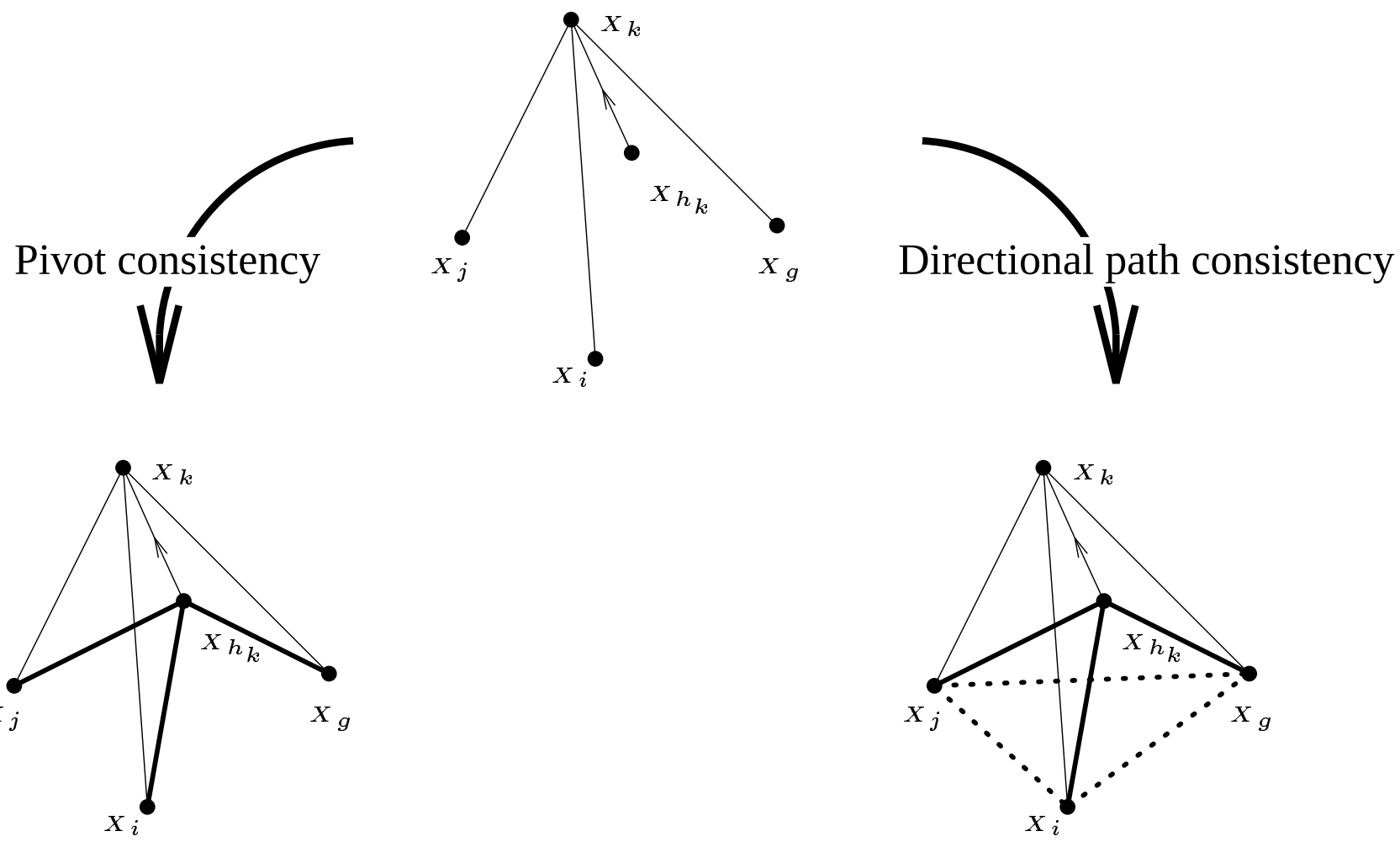

Figure 12: Processing pivot consistency and directional path consistency on node $x_k$, (with $g, h_k, i, j < k$): the bold edges represent the constraints possibly created by pivot consistency (w.r.t. $x_{h_k} \to x_k$), the dotted edges the extra ones possibly created by directional path consistency

Time complexity is lower for pivot consistency than for path consistency or directional path consistency, as will now be shown.

### 3.6 Complexity

Procedure `Pivot` is computed exactly once for each variable $x_k$ in $X - \mathcal{R}$; it then calls at most $k-1$ times procedure `Compatible` (once for each of the $k-1$ former variables). Achieving $x_k$-compatibility between two constraints $c_{h_k k}$ and $c_{ik}$ corresponds to achieving path consistency for relation $R_{hi}$ w.r.t. variable $x_k$. This operation needs $O(d^3)$ in general. But since the constraint $c_{h_k k}$ is functional, this complexity is now $O(d^2)$. The overall complexity of the filtering is therefore

$$O(\sum_{k=r+1}^{n} (k-1)d^2) = O((n^2 - r^2)d^2)$$

(instead of $O(n^3 d^3)$ for path consistency or directional path consistency[10]; $n$ is the number of variables, $d$ is the size of the domains and $r$ is the size of the root set).

## 4. From a Local Instantiation to a Solution

The intrinsic characteristic of a local consistency is to ensure that a partial instantiation can be extended to a new variable. We first present the properties of pivot consistency, in particular the conditions under which a consistent partial instantiation may be extended to a solution. We then explain how to compute the data required by the filtering algorithm, and finally present a method for solving functional CSPs.

### 4.1 Properties

We proceed in two stages: first, the addition of a new variable to the current instantiation, then the extension from the root set to a solution.

**Property 4.1** *Let $\mathcal{P} = (X, D, C, R)$ be a* CSP*, $I_{k-1}$ be a consistent instantiation of $Y_{k-1}$, and $x_k$ be a new variable to instantiate. If there exists $x_h \in Y_{k-1}$ such that $x_h \to x_k$ is a pivot of $Y_{k-1}$, then $I_{k-1}$ may be extended into $I_k$, a consistent instantiation of $Y_k$.*

**Proof:** Let us show that any constraint included in $Y_k$ is satisfied. Let $x_h \in Y_{k-1}$ such that $x_h \to x_k$ is a pivot of $Y_{k-1}$, and $a_h$ its value in $I_{k-1}$. Consequently, there exists $a_k = f_{h\to k}(a_h)$. Let us denote $I_k = (a_1, \ldots, a_h, \ldots, a_{k-1}, a_k)$.

1. Any constraint satisfied by $I_{k-1}$ and which does not contain $x_k$ is therefore obviously also satisfied by $I_k$.
2. For all $i < k$ s.t. $c_{ik} \in C$. First, $x_i \in Y_{k-1}$: let $a_i$ be its value in $I_{k-1}$. Second, $x_h \to x_k$ is a pivot of $Y_{k-1}$. $c_{ik}$ and $x_h \to x_k$ are consequently $x_k$-compatible. We thus have $(a_i, a_k) \in R_{ik}$; so, $c_{ik}$ is satisfied by $I_k$.

Every constraint included in $Y_k$ is satisfied by $I_k$. Hence $I_k$ is a consistent instantiation. □

10. Note that achieving directional path consistency on functional CSPs is still $O(n^3 d^3)$ and not $O(n^3 d^2)$, since first functional constraints are not necessarily directed the right way (where a functional constraint is "properly" directed if its origin is before its target in the instantiation ordering), and second non functional constraints may be created by the processing.

We now present the theorem at the center of the method we introduce further.

**Theorem 1** *Let $\mathcal{P} = (X, D, C, R)$, $\mathcal{R} \subseteq X$ a root set, and an $\mathcal{R}$-compatible assignment ordering such that $\mathcal{P}$ is pivot consistent w.r.t. them. If a consistent instantiation of $\mathcal{R}$ exists, then it can be extended to a backtrack-free solution.*

**Proof:** The variables of $X - \mathcal{R}$ only need to be instantiated along with the $\mathcal{R}$-compatible ordering: as each of them is the target of a pivot (since $\mathcal{P}$ is pivot consistent), Property 4.1 is verified step by step from the instantiation of $\mathcal{R}$ to a solution. □

### 4.2 A Solving-by-Decomposing Method

In section 4.1 (Theorem 1), we saw that, given a consistent instantiation of $\mathcal{R}$, and provided that the problem is pivot consistent w.r.t. an $\mathcal{R}$-compatible assignment ordering and a pivot set, we can not only guarantee there is a solution to the whole problem, but also find it without any backtracking. We can therefore deduce a method from this property. We now introduce this method, decomposed into four phases, and for each of them, we present the time complexity and the result of its computation on the example.

**Phase 1** Computation of a root set $\mathcal{R}$, a pivot candidate set $PC$ and an $\mathcal{R}$-compatible assignment ordering.

This phase can roughly be outlined as finding the sources of a graph where each strongly connected component is reduced to one node (computation of $\mathcal{R}$), finding a set of directed arcs covering all nodes in $X - \mathcal{R}$ ($PC$) and then performing topological sort (assignment ordering). For more details, see appendices A.1 and A.2.

*Complexity:* $O(e + n)$ to compute $\mathcal{R}$ (appendix A.1), and also $O(e + n)$ for $PC$ and the assignment ordering (appendix A.2). So, for the whole phase: $O(e + n)$.

The root set is $\mathcal{R} = \{$*GUIDES, CITIES*$\}$.

The pivot candidate set is $PC = \{$*CITIES* $\rightarrow$ *COUNTRIES*, *COUNTRIES* $\rightarrow$ *CURRENCIES*, *COUNTRIES* $\rightarrow$ *LANGUAGES*$\}$

The $\mathcal{R}$-compatible assignment ordering we chose is
*GUIDES, CITIES, COUNTRIES, CURRENCIES, LANGUAGES.*

**Phase 2** Pivot consistent filtering

*Complexity:* $O((n^2 - r^2)d^2))$ (see 3.6)

The pivot consistent problem is computed in section 3.4.

**Phase 3** Instantiation of $\mathcal{R}$

*Complexity:* $O(e_{\mathcal{R}} d^r)$, where $e_{\mathcal{R}}$ is the number of constraints included in $\mathcal{R}$.

There are five consistent instantiations of $\mathcal{R} = \{$*GUIDES, CITIES*$\}$:
$\{$*(Alice,Paris),(Bob,Paris),(Bob,London),(Bob,Washington),(Bob,New-York)*$\}$

**Phase 4** Instantiation of $X - \mathcal{R}$

*Complexity:* $O(n-r)$: extending a consistent instantiation of $\mathcal{R}$ to a solution only needs linear time.

The solutions of the problem are:

*(Alice,Paris,France,FrF,French)*
*(Bob,Paris,France,FrF,French)*
*(Bob,London,GBob,£,English)*
*(Bob, Washington, USA,$,English)*
*(Bob,New-York, USA,$,English)*

The overall time complexity of this method is therefore $O((n^2-r^2)d^2+e_{\mathcal{R}}d^r)$. This method thus reduces the search for one solution (or all) of a functional CSP to the one induced by its root set $\mathcal{R}$. As the time required obviously depends on the size of $\mathcal{R}$, we immediately see how interesting it is to compute a minimum sized root set. A direct consequence is that the smaller $\mathcal{R}$ is, the more efficient this method will be. It consequently seems worthwhile to know its size as soon as possible: this has be done in phase 1.

A last remark before concluding: the reader probably noticed that the number of consistent instantiations of $\mathcal{R}$ is the same as the number of solutions of the whole problem. This is not a coincidence, as we see below:

**Property 4.2** *Let $\mathcal{P}$ be pivot consistent, and let $\mathcal{R}$ be its root set. The number of solutions of $\mathcal{P}$ equals the number of consistent instantiations of $\mathcal{R}$: each solution is the extension of exactly one of those instantiations.*

**Proof:** Let $I_r = (a_1, \ldots, a_r)$ be a consistent instantiation of the root set $\mathcal{R} = \{x_1, \ldots, x_r\}$. According to Theorem 1, $I_r$ can be extended to at least one solution to $\mathcal{P}$. $\mathcal{R}$ is a root set of $\mathcal{G}$. Any $x_k$ in $X - \mathcal{R}$ is therefore a descendant from a variable $x_{r_k}$ of $\mathcal{R}$. There consequently exists a unique value $a_k \in D_k$ such that $(a_{r_k}, a_k) \in R_{r_k k}$, and, so, an only consistent instantiation of $\mathcal{R} \cup x_k$ including $I_r$, therefore only one solution including $I_r$. Moreover, as two different instantiations of $\mathcal{R}$ obviously cannot be extended to the same solution, there consequently exists a one to one mapping between the set of the consistent instantiations of $\mathcal{R}$ and the set $\mathcal{S}$ of the solutions to $\mathcal{P}$. □

## 5. Conclusion

Taking into account semantic properties of the constraints is a still recent approach to increase the efficiency of finding solutions to CSPs. This paper belongs to this frame.

We first introduced a new local consistency, pivot consistency, to deal with functional constraints for a lower cost than path consistency (we furthermore showed that pivot consistency is a weak form of path consistency). We then proposed an algorithm to achieve pivot consistent filtering. Later, we studied some properties that pivot consistency provides a functional CSP with, in particular conditions under which a consistent instantiation can be extended to a solution.

We were then led to present a decomposition method based on those properties, which decreases the complexity of solving functional CSPs. An interesting point is that the search for solutions to the former problem is reduced to the search for solutions to the subproblem induced by a particular subset, the root set; this new problem can be solved by any method, including heuristics.

Furthermore, we must add that this method deals with problems in which not all the constraints have to be functional (as we saw in the example). This is in our opinion an interesting improvement: most previous work dealing with properties of specific classes of constraints assumes that all the constraint of the network possess a given property, and only applies in this case.

However, some problems remain. First, and this is a classical problem as far as pre-processings are concerned: in what extent is the application of the method we present here useful? A partial answer can be given to this question: we indeed saw that one of the advantages of this method is that the cost of the search is known early in the process; one can thus estimate if the continuation of the process is worthwhile or not.

On the other hand, pivot consistency and the associated consistency properties can of course be generalized to non binary CSPs; yet, some problems appear: first of all, finding a minimum root set becomes exponential (it is the same problem as finding a *key* of minimum size in the field of relational databases, Lucchesi & Osborn, 1978); moreover, as we have seen, some constraints may be induced by the filtering processing; if this is not a problem for binary CSPs (the new constraints are still binary), in $n$-ary CSPs, some constraints may induce new ones of greater arity, which may from step to step lead to an explosion of the CSP arity.

Finding and characterizing classes of problems having root sets of small size as well as problems whose arity does not grow when achieving pivot consistency are therefore in the continuity of the work we presented in this paper.

To conclude, taking into account the semantics of constraints seems to be an interesting research area, and, as such, we can think of extending this kind of study (characterization of properties and of processings specific to a class of constraints) to other classes of constraints.

## Appendix A. Computation of Structural Data

### A.1 Computation of a Minimum Root Set $\mathcal{R}$

The example used in this paper is not suited to the illustration of the computation we present here; we therefore use another one (Figures 13 to 15).

***Computation of $\mathcal{G}_q$, reduced graph of $\mathcal{G}_f$:***

1: Compute $T = \{\tau_1, \dots, \tau_p\}$, the set of the strongly connected components of $\mathcal{G}_f$.

2: Compute graph $\mathcal{G}_q = (\mathcal{T}, C_q)$, where each vertex $t_i$ of $\mathcal{T}$ is the reduction of a strongly connected component $\tau_i$, and there exists an arc from $t_i$ to $t_j$ if and only if an arc existed from a vertex of $\tau_i$ to a vertex of $\tau_j$.

***Computation of the sources of $\mathcal{G}_q$, and the root set of $\mathcal{G}_f$:***

3: Compute the set $\mathcal{S}_q = \{t_{s_1}, \dots, t_{s_r}\}$ of the sources of $\mathcal{G}_q$.

4: Choose for each source $t_{s_k}$ of $\mathcal{S}_q$ any vertex of $\tau_{s_k}$, which will represent it.

5: The set $\mathcal{R}$ processed in that way is a root set of $\mathcal{G}_f$.

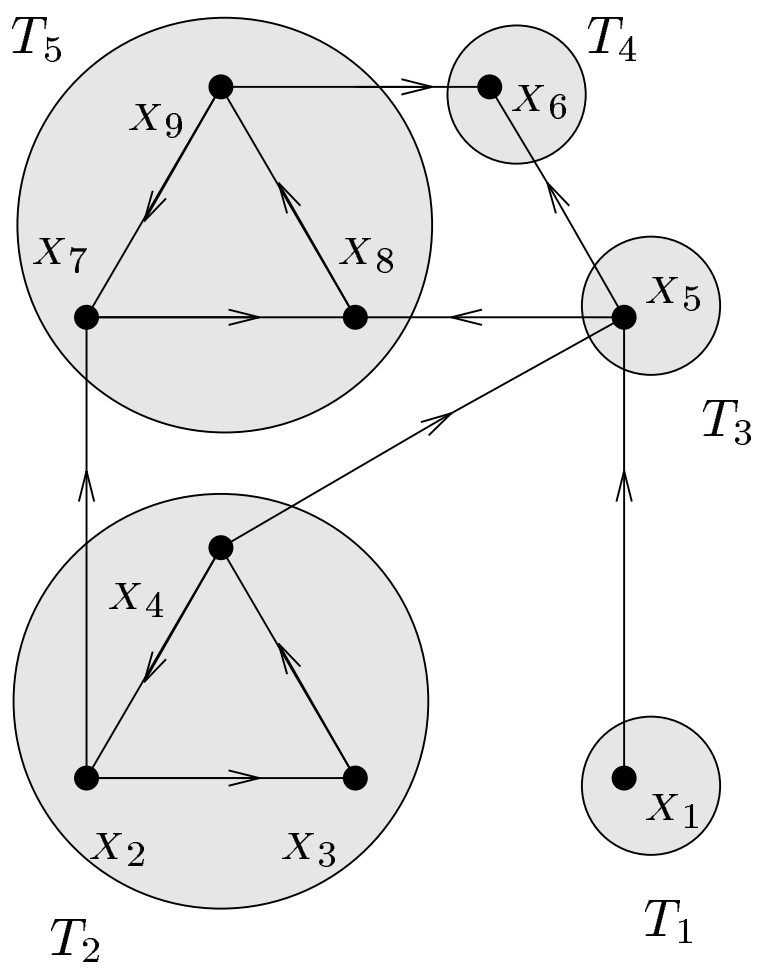

$T_1 = \{x_1\}$
$T_2 = \{x_2, x_3, x_4\}$
$T_3 = \{x_5\}$
$T_4 = \{x_6\}$
$T_5 = \{x_7, x_8, x_9\}$

Figure 13: Step 1: the strongly connected components

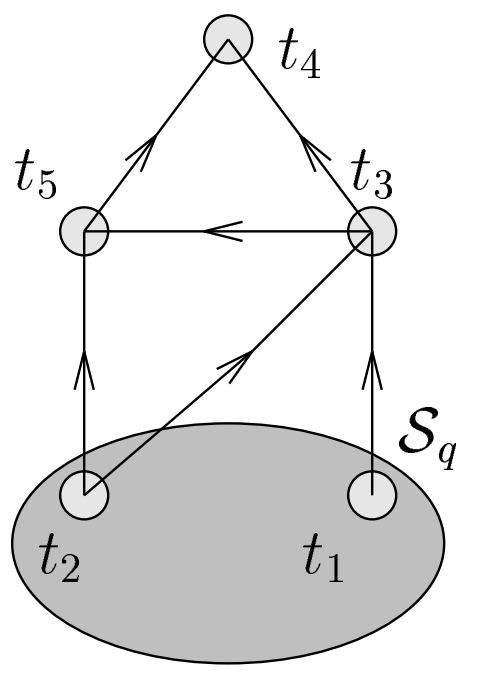

$\mathcal{S}_q = \{t_1, t_2\}$

Figure 14: Steps 2 and 3: the reduced graph and its sources

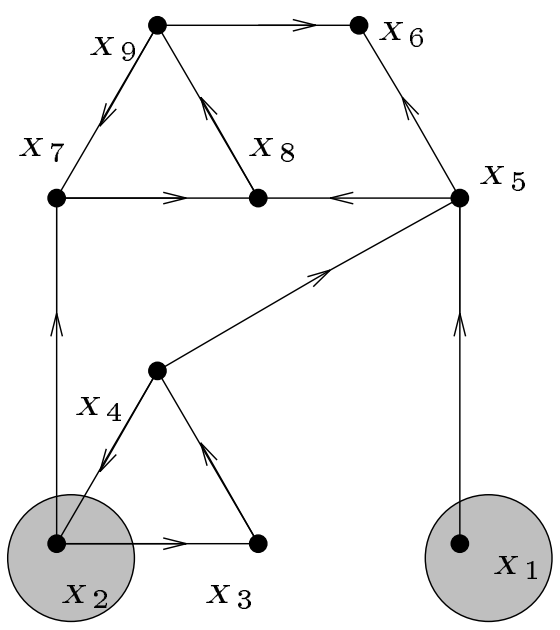

The only possible vertex to represent $t_1$ is $x_1$.
A vertex is chosen amongst $T_2 = \{x_2, x_3, x_4\}$ to represent $t_2$, for instance $x_2$.

We thus obtain $\mathcal{R} = \{x_1, x_2\}$

Figure 15: Steps 4 and 5: a minimum root set

A question arises: *Is the set $\mathcal{R}$ really a minimum root set*?
There are actually two questions: first, is $\mathcal{R}$ a root set? and if so, is this set *minimum*?

We first show that the set $\mathcal{R}$ we computed is a root set:
Let us show that any vertex is a descendant from an element of $\mathcal{R}$. Let $x_i$ be a vertex of $X - \mathcal{R}$. Two possibilities:

1. An element of $\mathcal{R}$ belongs to the same strongly connected component as $x_i$. By definition, $x_i$ is then a descendant from this element.

2. No element in $\mathcal{R}$ belongs to the same strongly connected component as $x_i$. Let us denote $\tau_i$ the strongly connected component which contains $x_i$, and $t_i$ its reduction in the graph $\mathcal{G}_q$. So, $t_i$ is not a source of this graph; consequently, there exists a source $t_{s_i}$ in $\mathcal{G}_q$ such that $t_i$ is a descendant from $t_{s_i}$. Let $x_{s_i}$ be the vertex of $X$ we chose to represent $t_{s_i}$ in $\mathcal{R}$. According to the definition of the reduced graph, knowing that $t_i$ is a descendant from $t_{s_i}$, any element of $\tau_i$ is a descendant from any element of $\tau_{s_i}$, and therefore from $x_{s_i}$. $x_i$ is consequently a descendant from an element of $\mathcal{R}$.

$\mathcal{R}$ is a root set. □

We now show that $\mathcal{R}$ is *minimum* (*i.e.*, there does not exist any root set of smaller size):
Let us denote $r = |\mathcal{R}|$. Assume there exists a root set $\mathcal{R}'$ whose size is $r' < r$. Let $SCC(\mathcal{R}')$ be the set of the Strongly Connected Components that contain all the elements of $\mathcal{R}'$, and let $\mathcal{T}(\mathcal{R}')$ be the set of their reductions in $\mathcal{G}_q$. By definition of $\mathcal{R}$, the reduced graph $\mathcal{G}_q$ has $r$ sources. There exists at least one source $t_{s_k}$ in $\mathcal{G}_q$ which does not belong to $\mathcal{T}(\mathcal{R}')$ (since $|\mathcal{T}(\mathcal{R}')| \leq r' < r$). Consequently, there exists no path from an element of $\mathcal{T}(\mathcal{R}')$ to $t_{s_k}$.
Let us denote $SCC(k)$ the strongly connected component reduced to $t_{s_k}$. From the definition of the reduced graph, there exists no path from an element of $\mathcal{R}'$ to an element of $SCC(k)$. Consequently, $\mathcal{R}'$ is not a root set.

$\mathcal{R}$ is thus a *minimum* root set. □

*Complexity:* It is the same as the complexity required for computing the strongly connected components, that is $O(e + n)$ (Tarjan, 1972), if $e$ is the number of edges and $n$ the number of vertices.

## A.2 Choice of the Pivots and Computation of an $\mathcal{R}$-compatible Order

We now describe how to choose the pivot candidates and to compute an $\mathcal{R}$-compatible ordering. We first present the conditions the pivot candidates must satisfy; we then present an algorithm that computes both $PC$ and an $\mathcal{R}$-compatible ordering. The definition of pivot consistency implies two conditions on the pivots:

1. Any $x_k$ in $X - \mathcal{R}$ must be the target of one and only one pivot, and no variable of $\mathcal{R}$ is the target of a pivot.

2. If $x_h \to x_k$ is a pivot, then $x_h$ is before $x_k$ in the ordering.

For each $x_k$ in $X - \mathcal{R}$, we have to choose a variable $x_h$ in $X$ so that $x_h \to x_k$ ($x_h$ necessarily exists, since $x_k$ does not belong to $\mathcal{R}$). Moreover, $PC$ must not contain any circuit, which would be in contradiction with condition 2. A way to prevent circuits is to mark every selected variable, and to select a new variable among the unmarked ones, which is the target of a functional constraint whose origin is already marked. This functional constraint is then included into $PC$, and its target (the new variable) is then marked. The set $PC$ we obtain this way is the set of the pivot candidates used by the pivot consistency algorithm (section 3.3). This set induces a partial order $\prec_{\mathcal{R}}$ on $X$ (we only have to add the transitivity constraints). In order to satisfy the conditions of Definition 3.3, the assignment ordering consequently has to be a linear extension of the partial order $\prec_{\mathcal{R}}$ prefixed by $\mathcal{R}$. Actually, computing $\prec_{\mathcal{R}}$ is not necessary: $PC$ is sufficient to compute the linear extension. There are no conditions on the variables of $\mathcal{R}$; the algorithm can thus be decomposed into two steps:

1. *Ordering of* $\mathcal{R}$
   Number from 1 to $r = |\mathcal{R}|$ the variables of $\mathcal{R}$ and mark them
2. *Ordering of* $X - \mathcal{R}$
   Repeat
      Choose a new unmarked variable $x_k$ in $X - \mathcal{R}$ s.t. there exists a marked $x_h$ s.t. $x_h \rightarrow x_k$
      Number and mark $x_k$
   Until all variables are marked

Figure 16 presents an algorithm computing both the set of pivot candidates $PC$ and an $\mathcal{R}$-compatible assignment ordering. Let us repeat that any linear extension issued from $PC$ is an $\mathcal{R}$-compatible ordering.

- The set `Marked` contains the variables that have already been treated
- The set `NextPossible` contains the next variables that can be chosen, that is the unmarked ones which are the target of a functional constraint whose origin is marked
- The sets `Origin[`$j$`]` contain the origins of functional constraints whose targets are described above (`NextPossible`)
- `Num` represents the number of the current variable in the assignment ordering

*Correctness:*

- Set $PC$:
  *Condition 1:* a functional constraint $x_h \rightarrow x_k$ is added to $PC$ each time a new variable $x_k$ is chosen: we therefore need to prove that any variable of $X - \mathcal{R}$ is selected *exactly* once, and that none is selected from $\mathcal{R}$. The variables are selected from `NextPossible`, which only contains unmarked variables; since no marked variable may be unmarked back, and every selected variable is marked, every variable can be selected at most once.

  Let us now prove that every variable of $X - \mathcal{R}$ is *actually* (at least once) selected: at any time, `NextPossible` is the set of all direct descendants of the marked variables that do not belong to $\mathcal{R}$; at each outer loop of step 2, a new variable is extracted from `NextPossible` and marked; from the definition of $\mathcal{R}$, all variables of $X - \mathcal{R}$ will therefore be reached and inserted in `NextPossible`.

```
begin
  Initializations
    Marked ← ∅
    NextPossible ← ∅
    for each i in 1..n do Origin[i] ← ∅
    Num ← 1
  Step 1: Numbering of R
    while Marked ≠ R do
      Choose x_i unMarked in R
      Mark x_i (* Add x_i to Marked *)
      Number x_i with Num
      Add 1 to Num
      for each x_i → x_j s.t.  x_j is unMarked and x_j ∉ R do
        Add x_j to NextPossible
        Add x_i to Origin[j]
      end for
    end while
  Step 2: Numbering of X − R — Construction of PC
    PC ← ∅
    while Marked ≠ X do
      Choose x_k in (and remove it from) NextPossible
      Mark x_k (* Add x_k to Marked *)
      Number x_k with Num
      Add 1 to Num
      Choose x_h in (and remove it from) Origin[k]
      Add x_h → x_k to PC
      for each x_k → x_p s.t.  x_p is unMarked do
        Add x_p to NextPossible
        Add x_k to Origin[p]
      end for
    end while
end
```

Figure 16: Computation of a set of pivot candidates $PC$ and an $\mathcal{R}$-compatible ordering

*Condition 2:* by construction of the sets `Origin[`$j$`]`, a variable $x_i$ is inserted in `Origin[`$j$`]` immediately after $x_i$ has been numbered; this only applies for the unmarked $x_j$. So, $x_j$ is necessarily numbered after $x_i$.

- Is this assignment ordering $\mathcal{R}$-compatible?
  The first variables are obviously those taken from $\mathcal{R}$ (step 1). The other condition is a direct consequence of condition 2 above.

*Complexity:*

Both steps have the same structure: choose a new variable $x_i$, mark and number it, and for each of its (unmarked) direct descendants, do some $O(1)$ operations (during step 2, a constraint is added to $PC$ which also requires $O(1)$). The overall time complexity is therefore

$$O(\sum_{i=1}^{n}(1 + |\Gamma^{+}(i)|)) = O(n + e)$$

where $\Gamma^{+}(i)$ is the set of the direct descendants from $x_i$.

## Acknowledgements

I wish to thank the "CONSTRAINTS" team of the Computer Science Laboratory of Montpellier for their helpful advice, and the anonymous reviewers for their useful comments and suggestions. I also thank Anne Bataller and Pascal Jappy for polishing the English. They all helped improve this paper. This research was partially supported by the PRC-IA research project "CSP FLEX" of the CNRS.